\documentclass[twoside,11pt]{article}

\usepackage{blindtext}

\usepackage[preprint]{jmlr2e}

\usepackage{amsmath,amsfonts,amssymb}
\usepackage{booktabs}
\usepackage{xcolor}
\usepackage[most]{tcolorbox}
\usepackage{subcaption}
\usepackage{hyperref}
\hypersetup{
  colorlinks=true,
  breaklinks=true,
  bookmarks=true,
  urlcolor=blue,
  citecolor=blue,
  linkcolor=blue,
  bookmarksopen=false,
  draft=false
}

\newcommand{\argmax}{\mathop{\rm argmax}}

\newtheorem{thm}{Theorem}[section]
\newtheorem{lem}{Lemma}[section]

\newtheorem{prop}{Proposition}[section]
\newtheorem{defn}{Definition}[section]

\def\approxcorrect{\checkmark\kern-1.1ex\raisebox{.89ex}{$\times$}}

\usepackage{amsmath,amsfonts,bm}

\def\eqref#1{equation~\ref{#1}}

\def\1{\bm{1}}

\DeclareMathAlphabet{\mathsfit}{\encodingdefault}{\sfdefault}{m}{sl}
\SetMathAlphabet{\mathsfit}{bold}{\encodingdefault}{\sfdefault}{bx}{n}

\def\gF{{\mathcal{F}}}

\def\gO{{\mathcal{O}}}

\newcommand{\E}{\mathbb{E}}

\newcommand{\R}{\mathbb{R}}

\newtheorem{assumption}{Assumption}

\ShortHeadings{Beyond EIG: Stable BOED with IPMs}{Wu, Liang and Yang}
\firstpageno{1}

\begin{document}

\title{Beyond Expected Information Gain: Stable Bayesian Optimal Experimental Design with Integral Probability Metrics and Plug-and-Play Extensions}

\renewcommand{\thefootnote}{\fnsymbol{footnote}}

\author{\name Di Wu\footnotemark[1] \email dwuwd@umd.edu \\
       \addr Department of Mathematics\\
       University of Maryland, College Park\\
       MD 20742, USA
       \AND
       \name Ling Liang\footnotemark[1] \email liang.ling@u.nus.edu \\
       \addr Department of Mathematics\\
       The University of Tennessee, Knoxville\\
       TN 37916, USA
       \AND
       \name Haizhao Yang\footnotemark[2] \email hzyang@umd.edu \\
       \addr Department of Mathematics and Department of Computer Science\\
       University of Maryland, College Park\\
       MD 20742, USA}

\footnotetext[1]{Equal contribution.}
\footnotetext[2]{Corresponding author.}
\renewcommand{\thefootnote}{\arabic{footnote}}

\editor{My editor}

\maketitle

\begin{abstract}
Bayesian Optimal Experimental Design (BOED) provides a rigorous framework for decision-making tasks in which data acquisition is often the critical bottleneck, especially in resource-constrained settings. Traditionally, BOED typically selects designs by maximizing expected information gain (EIG), commonly defined through the Kullback-Leibler (KL) divergence. However, classical evaluation of EIG often involves challenging nested expectations, and even advanced variational methods leave the underlying log-density-ratio objective unchanged. As a result, support mismatch, tail underestimation, and rare-event sensitivity remain intrinsic concerns for KL-based BOED. To address these fundamental bottlenecks, we introduce an IPM-based BOED framework that replaces density-based divergences with integral probability metrics (IPMs), including the Wasserstein distance, Maximum Mean Discrepancy, and Energy Distance, resulting in a highly flexible plug-and-play BOED framework. We establish theoretical guarantees showing that IPM-based utilities provide stronger geometry-aware stability under surrogate-model error and prior misspecification than classical EIG-based utilities. We also validate the proposed framework empirically, demonstrating that IPM-based designs yield highly concentrated credible sets. Furthermore, by extending the same sample-based BOED template in a plug-and-play manner to geometry-aware discrepancies beyond the IPM class, illustrated by a neural optimal transport estimator, we achieve accurate optimal designs in high-dimensional settings where conventional nested Monte Carlo estimators and advanced variational methods fail.

\end{abstract}
\vspace{1em}

\section{Introduction}
Optimal Experimental Design (OED) provides a principled framework for selecting experiments to learn unknown model parameters as efficiently as possible \citep{steinberg1984experimental}. It has been widely used across science and engineering, including parameter inference in scientific computing, such as partial differential equations \citep{huan2013simulation,hellmuth2025data}, and inverse problems \citep{haber2008numerical,alexanderian2016fast,ruthotto2018optimal,alexanderian2021optimal,helin2022edge,wu2023fast,liang2024pnod,chang2025provable,chen2026data}. In many of these applications, data acquisition is a major bottleneck because experiments can be expensive, simulations may require substantial computational resources, and only a limited number of measurements can be collected. This challenge arises in a wide range of settings including materials discovery \citep{lookman2019active,chen2021optimal}, calibration of expensive computational simulators \citep{sacks1989design,dunbar2022ensemble,sun2024arma}, and data selection and training in modern machine learning \citep{chen2015selection,yang2016computational,fiez2019sequential,camilleri2021high,wagenmaker2022instance,bhatt2024experimental,huadaptive,zhu2025balancing,gao2026observation}. In such resource-constrained regimes, one must carefully decide which experiments to perform in order to maximize the information gained about the underlying system. In this context, the use of prior knowledge plays an essential role.

Bayesian Optimal Experimental Design (BOED) extends OED by explicitly incorporating prior uncertainty about the unknown parameters \citep{lindley1956measure}. Let $x \in \mathcal{X}$ denote the unknown model parameters, $\Xi$ the design domain, $\xi \in \Xi$ an experimental design, and $y \in \mathcal{Y}$ the corresponding experimental outcome with $\mathcal{Y}$ being the space of all possible outcomes. Given a prior distribution $p(x)$, classical BOED seeks a design $\xi^*$ that maximizes the expected information gain (EIG), commonly defined through the Kullback--Leibler (KL) divergence from the posterior to the prior \citep{shannon1948mathematical}:
\begin{equation*}
\xi^* := \argmax_{\xi \in \Xi}
\left\{
\mathrm{EIG}(\xi)
:=
\int_{\mathcal{Y}} \int_{\mathcal{X}}
\log\!\left(\frac{p(x \mid y,\xi)}{p(x)}\right)
\, p(x \mid y,\xi)\, p(y \mid \xi)\, dx\, dy
\right\}.
\end{equation*}

Despite its conceptual appeal, the classical EIG formulation raises significant challenges. From a computational perspective, evaluating EIG requires estimating nested expectations, which is typically expensive and difficult to do accurately \citep{rainforth2018nesting}, and standard Monte Carlo methods often suffer from high variance and slow convergence. From a modeling perspective, the main issue is structural rather than estimator-specific. Even when advanced estimators \citep{foster2019variational} reduce the computational burden of EIG evaluation, the utility remains a log-density-ratio functional. Consequently, it is inherently sensitive to tail underestimation, near-zero likelihood regions, and rare but highly informative observations. Moreover, modern BOED pipelines often replace continuous priors or posteriors with empirical approximations, and the KL divergence is not always a natural perturbation metric between such exact and approximate measures because the required absolute continuity may fail.

To address these limitations, we propose a geometry-aware reformulation of BOED based on Integral Probability Metrics (IPMs) \citep{muller1997integral}. IPMs, including the Wasserstein distance \citep{dudley2018real}, Maximum Mean Discrepancy (MMD) \citep{gretton2009fast,berlinet2011reproducing,gretton2012kernel}, and Energy Distance \citep{szekely2004testing,szekely2005new}, quantify the discrepancy between two distributions through their action on a prescribed class of test functions. We introduce an \textbf{IPM-based Bayesian Optimal Experimental Design} framework by defining the design utility as

\begin{equation*}
    \max_{\xi \in \Xi} U_{\mathcal{F}}(\xi) := \int_{\mathcal{Y}} \gamma_{\mathcal{F}}\big(p(x), p(x \mid y, \xi)\big) p(y \mid \xi) dy,
\end{equation*}
where $\gamma_{\mathcal{F}}$ is the IPM induced by a function class $\mathcal{F}$. This formulation is naturally flexible such that the same sample-based BOED template can also be extended in a plug-and-play manner to other discrepancies based on domain knowledge.

This IPM-based framework offers both theoretical and computational advantages over the traditional EIG-based formulation. On the theoretical side, we conduct a systematic stability analysis to characterize how surrogate-model and prior errors affect BOED utilities. In particular, we show that bounded-kernel MMD, as a bounded IPM, admits universal stability bounds controlled directly by the surrogate likelihood error or the prior perturbation, without requiring strict tail assumptions. We further analyze the unbounded IPMs considered in this paper, namely Energy Distance and the 1-Wasserstein distance, and clarify the tail and smoothness conditions required for stability. By contrast, the classical KL divergence does not admit comparable robustness guarantees because of its dependence on density ratios. On the computational side, IPMs enable stable sample-based utility estimation without explicit density-ratio evaluation, leading to substantial efficiency gains. Moreover, in our experiments, IPM-based utilities exhibit broader near-optimal design regions and more stable optimization landscapes than KL-based criteria. More generally, BOED can be built from any discrepancy between the prior and posterior. In this work, our theoretical development focuses on IPM utilities because they admit a unified stability analysis. At the same time, the resulting sample-based BOED pipeline is modular, which allows us to incorporate geometry-aware discrepancies beyond the IPM class at the computational level. We illustrate this extension by incorporating a neural optimal transport estimator \citep{amos2017input} in high-dimensional settings.

\paragraph{Contributions.} The main contributions of this work are summarized as follows:
\begin{itemize}
    \item \textbf{A stable BOED framework based on Integral Probability Metrics.}
    We develop a BOED formulation in which the design utility is defined through an Integral Probability Metric between the prior and posterior distributions. This provides a geometry-aware alternative to classical KL-based expected information gain and avoids direct dependence on density ratios. Within this framework, we establish utility perturbation bounds with respect to both surrogate likelihood error and prior misspecification, and we make explicit the different assumptions required for bounded and unbounded IPMs.
    \item \textbf{Reliable sample-based utility estimation and improved optimization behavior.}
    We show empirically that IPM-based utilities admit stable sample-based estimation without explicit density estimation or log-ratio evaluation. In our experiments, this leads to substantial computational advantages over KL-based criteria, while also producing broader near-optimal design regions and more reliable optimization landscapes.
    \item \textbf{Plug-and-play extensions beyond the IPM class.}
    Beyond the IPM theory developed in this paper, we show that the same sample-based BOED pipeline is modular enough to incorporate other geometry-aware discrepancies. As a representative example, we integrate a neural optimal transport estimator for the squared $2$-Wasserstein distance in high-dimensional settings where nested Monte Carlo approaches and variational methods become computationally prohibitive or unreliable, illustrating that the proposed computational framework extends naturally beyond the IPM class.
\end{itemize}

The rest of this paper is organized as follows. Section \ref{sec:related_work} reviews the relevant literature. Section \ref{sec:ipms} provides the formal definition of IPMs, together with numerical results that motivate our approach. In Section \ref{sec:stability_analysis}, we establish the theoretical guarantees for the proposed framework and provide a detailed discussion. Section \ref{sec:experiments} presents a series of experiments that validate the robustness and demonstrate the scalability and flexibility of the plug-and-play framework. Finally, Section \ref{sec:conclusion} concludes the paper with a summary and directions for future research. All formal proofs and supplementary mathematical details are deferred to the appendix.

\section{Related Work}
\label{sec:related_work}
\paragraph{Classical Bayesian Optimal Experimental Design and Expected Information Gain.} 
Classical BOED relies on an information-theoretic strategy, where Shannon entropy \citep{shannon1948mathematical} forms the basis for the expected information gain (EIG) \citep{lindley1956measure, bernardo1979expected, shewry1987maximum, sebastiani2000maximum}. Expected Fisher information has also been studied as an alternative optimality criterion \citep{walker2016bayesian, overstall2022properties, prangle2023bayesian}. Despite its elegant theoretical foundation, estimating the EIG is difficult in practice because it involves nested expectations, which can be computationally expensive and statistically unstable. To overcome this computational bottleneck, various estimation strategies have been developed, including nested Laplace approximations \citep{long2013fast, beck2018fast} and nested Monte Carlo estimators \citep{myung2013tutorial, rainforth2018nesting}. To further improve efficiency, recent variational approaches utilize function approximation to relax the computational burden imposed by strict nesting \citep{barber2004algorithm,huan2013simulation, kleinegesse2019efficient, foster2019variational, foster2021deep,jin2024continuous,ohn2024adaptive}. For broader discussions of the development of BOED, we refer the reader to several surveys \citep{chaloner1995bayesian, ryan2016review, rainforth2024modern, huan2024optimal}.

Our goal is not to claim that KL-based BOED is always computationally infeasible. Rather, we stress that estimator-level improvements do not change the underlying stability properties of the KL objective. In contrast, our work changes the utility itself and studies stability at the level of probability measures.

\paragraph{Optimal Transport and Wasserstein Information Criteria.}
While KL divergence is widely used in BOED, it is fragile when dealing with non-overlapping supports \citep{arjovsky2017wasserstein}. To address this, recent concurrent works \citep{helin2025bayesian,kerrigan2025geometric} have explored using optimal transport \citep{peyre2019computational,cang2022supervised,wu2025pins} and the Wasserstein distance \citep{villani2009optimal,liu2021wasserstein,song2026wasserstein} as measures between probability distributions, which is often more natural when supports are poorly aligned or when empirical particle approximations are used. This line of work provides an important step toward geometry-aware BOED. Our work is closely related in spirit, but differs in scope: rather than focusing on a single transport criterion, we develop a unified BOED view based on Integral Probability Metrics, which allows us to compare several geometry-aware discrepancies within one theoretical framework.

\paragraph{IPMs and Neural Discrepancy Estimators.}
Integral Probability Metrics (IPMs) \citep{muller1997integral} have gained significant attention in machine learning as robust, geometry-aware alternatives to density-based divergences. Recent studies \citep{amos2017input,korotin2019wasserstein,makkuva2020optimal,birrell2022f} also explore learning-based discrepancy estimation. These methods provide scalable sample-based tools for comparing probability measures in high-dimensional settings by parameterizing transport potentials, maps, or dual objectives with neural networks. Our use of OT-ICNN in the high-dimensional experiments should be understood in this computational sense. The theory developed in this paper focuses on IPM-based BOED utilities, while the sample-based BOED pipeline is modular enough to accommodate geometry-aware discrepancies beyond the IPM class.

\section{Integral Probability Metrics for BOED}
\label{sec:ipms}

To construct a stable utility function for BOED, we replace the standard Kullback-Leibler (KL) divergence with Integral Probability Metrics (IPMs). In this section, we formally define the IPM framework, introduce the three IPMs that form the theoretical core of the paper, and illustrate the fundamental fragility of KL divergence in rare-event regimes.

\subsection{Integral Probability Metrics}
Integral probability metrics are a family of distances between two probability distributions defined by the maximum difference in their expectations over a specified class of test functions. Let $P$ and $Q$ be probability measures on the measurable parameter space $\mathcal{X}$, and let $\mathcal{F}$ be a set of measurable real-valued functions $f: \mathcal{X} \to \mathbb{R}$. The IPM induced by the function class $\mathcal{F}$ is defined as:
\begin{equation*}
    \gamma_{\mathcal{F}}(P, Q) = \sup_{f \in \mathcal{F}} \big| \mathbb{E}_{X \sim P}[f(X)] - \mathbb{E}_{Y \sim Q}[f(Y)] \big|.
\end{equation*}
By restricting the complexity of the test function class $\mathcal{F}$, IPMs offer a geometry-aware measure of distance that does not require the distributions to have overlapping support or well-behaved density ratios. In this work, we focus on three representative IPMs characterized by distinct function classes:
\begin{enumerate}
    \item \textbf{1-Wasserstein Distance ($W_1$):} By the Kantorovich--Rubinstein duality theory \citep{hanin1992kantorovich}, the 1-Wasserstein distance is defined as 
    \[W_1(P,Q)=\sup_{\|f\|_L\leq 1} \big|\mathbb{E}_{X \sim P}[f(X)] - \mathbb{E}_{Y \sim Q}[f(Y)]\big| \]
    where $\|\cdot\|_L$ denotes the Lipschitz constant. Thus, \(W_1\) is the IPM induced by the class of \(1\)-Lipschitz functions;
    \item \textbf{Maximum Mean Discrepancy (MMD):} Let \(\mathcal{H}_k\) be a reproducing kernel Hilbert space (RKHS) \citep{smola2007hilbert} with a kernel \(k\). The corresponding MMD is
    \[\mathrm{MMD}_k(P,Q)=\sup_{\|f\|_{\mathcal{H}_k}\leq 1} \big|\mathbb{E}_{X \sim P}[f(X)] - \mathbb{E}_{Y \sim Q}[f(Y)]\big|.\]
    Hence, MMD is the IPM induced by the unit ball of the RKHS \(\mathcal{H}_k\);
    \item \textbf{Energy Distance (ED):} Let \(\rho:\mathcal{X}\times\mathcal{X}\to\mathbb{R}\) be a semimetric of negative type. The associated energy distance is
    \[
    \mathrm{ED}_{\rho}(P,Q)
    =
    \left(
    2\mathbb{E}\rho(X,Y)
    -
    \mathbb{E}\rho(X,X')
    -
    \mathbb{E}\rho(Y,Y')
    \right)^{1/2},
    \]
    where \(X,X'\sim P\) and \(Y,Y'\sim Q\) are independent, respectively. As shown by \citet{sejdinovic2013equivalence}, if we set $k_{x_0}(x,y)=\frac{1}{2}\bigl(\rho(x,x_0)+\rho(y,x_0)-\rho(x,y)\bigr)$ for a fixed $ x_0\in\mathcal{X},
    $
    then \(k_{x_0}\) is a positive definite kernel and
    \[
    \mathrm{ED}_{\rho}(P,Q)
    =
    \sqrt{2}\,\mathrm{MMD}_{k_{x_0}}(P,Q).
    \]
    Therefore, energy distance is equivalent to an IPM induced by the unit ball of the RKHS associated with \(k_{x_0}\).
\end{enumerate}

While Energy Distance (ED) can be viewed as a special case of MMD, the two exhibit significant differences in both theory and practice. As we will discuss in Section \ref{sec:stability_analysis}, the standard configurations including MMD with bounded kernels and ED with Euclidean distance lead to distinct theoretical guarantees. Practically, MMD offers greater flexibility through various kernel choices, whereas ED is often more robust. Note that the high-dimensional experiments in Section \ref{sec:highd_exp} should be viewed as plug-and-play extensions beyond the IPM class and are not covered by the theory.

\subsection{Motivating Example: The Fragility of KL Divergence}
\begin{figure}[h]
  \centering
  \includegraphics[width=\textwidth]{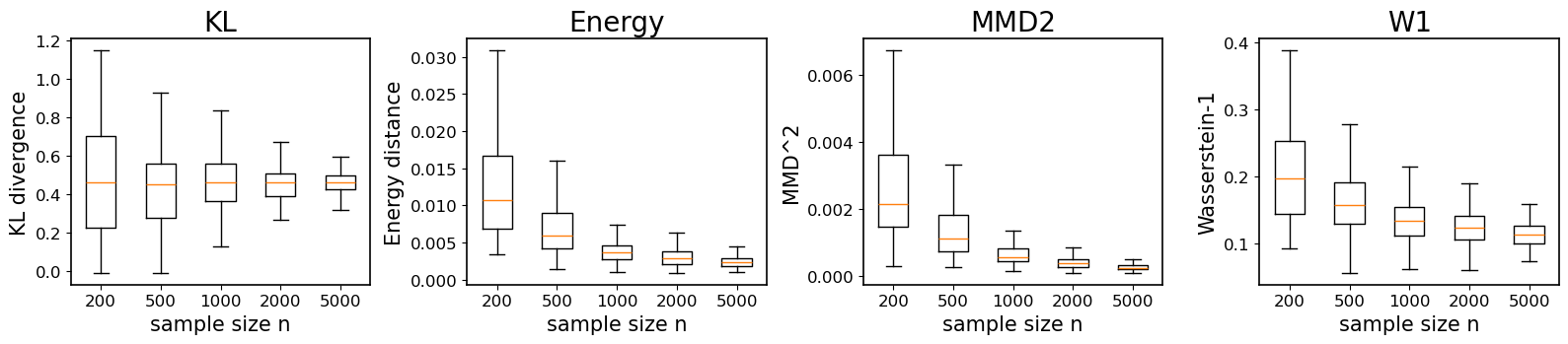} 
    \vspace{0.2cm}
    \includegraphics[width=\textwidth]{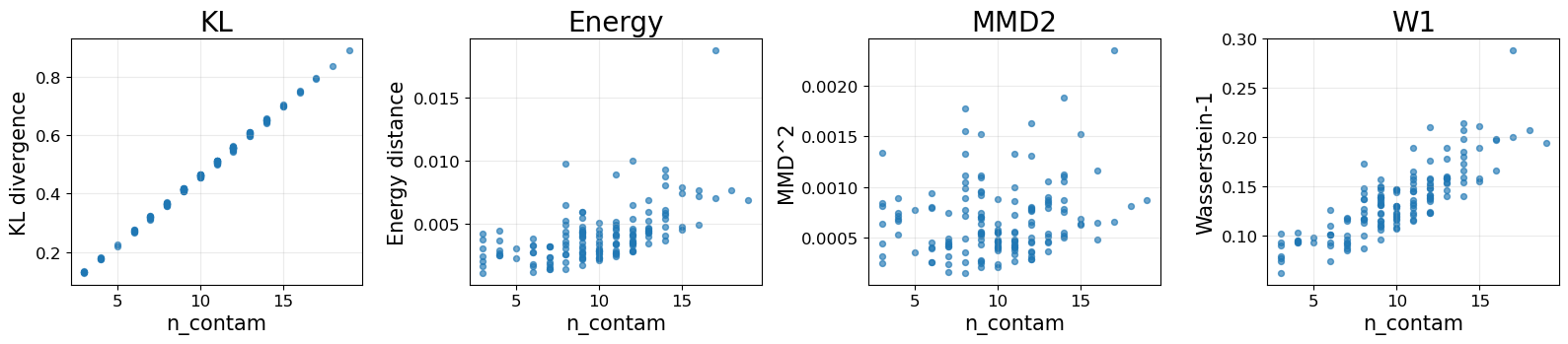}
  \caption{Rare-event contamination experiment with $\epsilon = 0.01$. \textbf{Top row:} Distributions of sample-based distance estimates across trials as the sample size $n$ increases. \textbf{Bottom row:} How each estimator responds to the actual contamination count $n_{contam}$ in the $P$ batch for a fixed sample size $n = 1000$. The KL divergence is strongly sensitive to $n_{contam}$ (showing an almost linear dependence), indicating extreme variance driven by rare samples. In contrast, the empirical $W_1$ shows a stable two-sample noise floor that predictably decreases with $n$, while ED and MMD vary smoothly and robustly with $n_{contam}$.}
  \label{fig:contam}
\end{figure}
To demonstrate where KL-based measures typically fail in practice, we compare these sample-based estimators using a controlled rare-event contamination setup. Consider two distributions that differ only by a low-probability component:
\begin{equation*}
    Q \sim \mathcal{N}(0, 1), \quad P = (1-\epsilon)\mathcal{N}(0, 1) + \epsilon\mathcal{N}(10, 0.1^2), \quad \text{with } \epsilon = 10^{-2}.
\end{equation*}
Here, the difference between $P$ and $Q$ comes down to tail events, which rarely appear consistently in small sample batches. 

The bottom row of Figure \ref{fig:contam} shows that all discrepancies respond to the presence of contaminated samples, but they do so in qualitatively different ways. The KL estimate is almost locked to the exact contamination count, producing an approximately linear response and hence large trial-to-trial variance. By contrast, the IPM estimates are influenced by the contaminated samples through pairwise geometry or transport cost, which leads to a smoother and less count-dominated dependence. In particular, ED and $W_1$ show a clear but gradual increase with $n_{\mathrm{contam}}$, while MMD exhibits a noisier finite-sample response due to kernel interactions. The key point is therefore not that IPMs are insensitive to rare events, but that they avoid the near-deterministic count domination exhibited by the KL estimate.

This phenomenon is relevant to BOED because many designs are evaluated under observations for which most samples are weakly informative, while a small subset of rare observations can induce much larger posterior shifts. In such settings, estimators built on ratio-type quantities may become highly unstable unless additional structure or variance-reduction machinery is introduced. This motivates our subsequent focus on utility formulations whose perturbation behavior can be controlled directly at the measure level.

\section{Stability Analysis of IPMs}
\label{sec:stability_analysis}

Motivated by \cite{helin2025bayesian}, we ask the following question regarding utility stability:
\begin{tcolorbox}[colback=white,colframe=black,boxrule=0.8pt,arc=2pt]
\begin{center}
\textit{If one slightly perturbs the mathematical model (i.e., the likelihood) or the initial assumptions (i.e., the prior), how much does the design utility change?}
\end{center}
\end{tcolorbox}

As shown in Section \ref{sec:ipms}, the severe instability of the KL divergence in rare-event regimes motivates the paradigm shift toward IPMs. Our theoretical framework extends the recent work of \cite{helin2025bayesian}, who established the use of the Wasserstein distance as a valid, geometrically meaningful OED criterion, while we adopt a broader IPM framework. A key advantage of these metrics is that they avoid density ratios entirely, thereby circumventing the tail sensitivity that limits standard EIG. Within this unified IPM framework, our theory highlights a fundamental theoretical trade-off between geometric richness and mathematical regularity.

For MMD, we primarily consider the most commonly-used setting, translation-invariant radial kernels \citep{tolstikhin2017minimax}, such as Gaussian, Mixture of Gaussians, Inverse Multiquadrics, and Mat\'ern kernels. For these kernels, the test functions are strictly bounded. By the reproducing property, we have
$$\sup_{\|f\|_{\mathcal{H}_k}\leq 1}|f(x)|=\sup_{\|f\|_{\mathcal{H}_k}\leq 1}|\langle f,k(x,\cdot)\rangle|\leq\sup_{\|f\|_{\mathcal{H}_k}\leq 1}\|f\|_{\mathcal{H}_k}\sqrt{k(x,x)}=O(1),$$
where the final equality follows directly from transition invariance. Consequently, MMD acts as a bounded IPM in the following analysis. 

In contrast, ED is generally unbounded. For the standard Euclidean energy distance, the growth rate of the test functions is $$\sup_{\|f\|_{\mathcal{H}_{k_{x_0}}}\leq 1}|f(x)|\leq\sqrt{\rho(x,x_0)}=\gO(\|x\|^{1/2}).$$
The 1-Wasserstein distance exhibits even faster growth, with an envelope of $\mathcal{O}(\lVert x \rVert)$. To guarantee utility stability when using unbounded IPMs like ED and Wasserstein, stricter mathematical conditions are required: for instance, the prior exhibits sub-Gaussian tails, and the forward model is globally Lipschitz continuous.

\subsection{Mathematical Setup and Notation} 
We begin by formalizing our mathematical setting. Let $\mathcal{X}$ and $\mathcal{Y}$ denote the parameter and observation spaces, respectively. Because we consider a fixed experimental design throughout this section, we streamline our notation by leaving the design condition implicit within both the forward model and the resulting likelihood. For a given parameter $x \in \mathcal{X}$, we define the forward model $G: \mathcal{X} \to \mathcal{Y}$ as a mapping to the observation space. We denote the class of all such potential forward mappings as $\mathcal{G}$, where $G \in \mathcal{G}$. The model generates data $y \in \mathcal{Y}$ according to $y = G(x) + \eta$, where $\eta$ represents observational noise, thereby inducing the likelihood $p(y|x)$. With a slight abuse of notation, for a given prior distribution $\mu(dx)$, the evidence (i.e., marginal likelihood) is given by $p(y) = \int_{\mathcal{X}} p(y|x)\mu(dx)$, yielding the posterior distribution $\mu^y(dx) = \frac{p(y|x)}{p(y)}\mu(dx)$.

Building on the IPM framework from Section \ref{sec:ipms}, we define the expected utility of an experimental design based on the discrepancy between the prior and posterior. To guarantee that the IPM $\gamma_{\mathcal{F}}$ behaves as a true distance metric—specifically, satisfying symmetry such that $\gamma_{\mathcal{F}}(P, Q) = \gamma_{\mathcal{F}}(Q, P)$, the measurable function class $\mathcal{F}$ must be symmetric (i.e., $f \in \mathcal{F} \Rightarrow -f \in \mathcal{F}$). This property holds natively for many standard function classes, including those defining the Wasserstein distance, MMD, and energy distance. The expected IPM utility for a given experimental design is therefore formulated as:
$$U_{\mathcal{F}} = \mathbb{E}_{p(y)} \left[ \gamma_{\mathcal{F}}(\mu, \mu^y) \right] = \int_{\mathcal{Y}} \gamma_{\mathcal{F}}(\mu, \mu^y)p(y)dy.$$

\subsection{Likelihood Stability}
\label{sec:likelihood_stability}
In computationally intensive scientific applications, the true forward model $G(x)$ is frequently replaced by a cheaper surrogate, such as a neural network. Because these surrogates are only approximations, they inevitably introduce localized errors into the likelihood function \citep{liu2023review}. A robust experimental design framework must be resilient to these inaccuracies, meaning its utility degrades smoothly. As illustrated in our motivating example, the traditional EIG fails this criterion since a minor surrogate error in a low-probability region can cause the underlying KL divergence to explode logarithmically.

To rigorously demonstrate how IPMs resolve this fragility and maintain stability under surrogate modeling errors, we shall first quantify the ``topological severity'' of the metric's test functions. We capture this via a pointwise growth profile.

\begin{defn}[Growth Profile]
    For an IPM defined by a symmetric function class $\mathcal{F}$,  its pointwise growth profile is defined as $\omega_{\mathcal{F}}(x) := \sup_{f \in \mathcal{F}} |f(x)|$. The expected growth under the prior $\mu$ is denoted as $\bar{\omega} = \int_{\mathcal{X}} \omega_{\mathcal{F}}(x)\mu(dx)$. 
\end{defn}

Essentially, the growth profile characterizes how severely the metric penalizes large deviations. As discussed in the beginning of this section, for bounded IPMs such as MMD with standard radial kernels, the growth profile is strictly bounded: $\omega_{\mathcal{F}}(x) \le M < \infty$. For unbounded metrics, we have $\omega_{\mathcal{F}}(x) = \mathcal{O}(\|x\|^{1/2})$ for energy distance with standard Euclidean distance, and $\omega_{\mathcal{F}}(x) = \mathcal{O}(\|x\|)$ for the 1-Wasserstein distance.

\begin{thm}[Likelihood Stability Bound] \label{thm:likelihood}
Let two forward models yield likelihoods $p(y|x)$ and $p_*(y|x)$, evidences $p(y)$ and $p_*(y)$, and expected utilities $U_{\gF}$ and $U_{\gF}^*$, respectively. Then, the difference in their utility is bounded by the expected $L_1$ error of the likelihoods, weighted by the growth profile:
$$|U_{\mathcal{F}} - U_{\mathcal{F}}^*| \le \int_{\mathcal{Y}} \int_{\mathcal{X}} \Big( \omega_{\mathcal{F}}(x) + \bar{\omega} \Big) \Big| p(y|x) - p_*(y|x) \Big| \mu(dx) dy.$$
\end{thm}
\begin{proof}
    See Appendix \ref{sec:appendix_likelihood_proof}.
\end{proof}

The bound presented in Theorem \ref{thm:likelihood} mathematically decouples the inherent topological properties of the metric from the approximation error of the surrogate model. Specifically, the utility perturbation is controlled linearly by the expected $L_1$ error of the likelihoods, weighted by the growth profile $\omega_{\mathcal{F}}(x)$. This reveals a fundamental mechanism for stability where the penalty incurred by a surrogate's error is strictly affected by how fast the metric's chosen test functions are allowed to grow.

Furthermore, our result clearly captures the distinct behaviors of bounded and unbounded metrics. For bounded IPMs, $\omega_{\mathcal{F}}(x)$ is globally capped, providing unconditional and uniform $L_1$ control over utility errors. This guarantees that utility degrades gracefully alongside surrogate accuracy. Conversely, for unbounded IPMs, the growth of $\omega_{\mathcal{F}}(x)$ indicates that surrogate inaccuracies in the tails of the parameter space incur a higher penalty. This establishes a clear theoretical trade-off between the rich geometric transport of unbounded metrics and the universal stability guarantees of bounded ones. We provide a deeper investigation into the behavior of $\omega_{\mathcal{F}}(x)$ and its practical implications in Section \ref{sec:understanding}.

\paragraph{Contrast with KL Divergence.}  To fully understand the robustness of IPMs, it is helpful to compare them with standard information-theoretic approaches. Consider the KL divergence, namely, $D_{KL}(P\|Q) := \int p(x) \log\left(\frac{p(x)}{q(x)}\right) dx$, which implicitly utilizes the log-ratio as its test function. Because EIG relies on KL divergence, it scales with the logarithm of the likelihood ratio and reacts to multiplicative errors rather than additive $L_1$ differences. Consequently, its stability cannot be neatly factored into a bounded growth profile. If a surrogate assigns near-zero likelihood where the true model assigns a non-negligible likelihood, the KL divergence blows up. This clearly demonstrates why EIG is highly unstable to even minor misspecifications in the tails of surrogate models. IPMs remove the ratio-based amplification mechanism of KL. For bounded IPMs this yields uniform control, while for unbounded IPMs rare observations may still matter, but their effect is governed by kernel geometry and moment conditions rather than logarithmic small-denominator instability.


\subsection{Prior Stability}
While Section \ref{sec:likelihood_stability} establishes robustness against errors in the forward model, practical BOED faces another critical source of approximation error from the prior distribution. In most standard computational setups, the continuous prior $\mu$ is replaced by a discrete empirical measure $\tilde{\mu}$ using Monte Carlo sampling. 

To bound the resulting utility error, we exploit the shift-invariance geometric property shared by standard IPMs. The following lemma demonstrates that we can always anchor our test functions to the approximate prior without changing the underlying distance.

\begin{lem}[Prior-Anchored Supremum] \label{lem:anchor}
    Let $\mathcal{F}$ be a symmetric function class defining an IPM $\gamma_{\mathcal{F}}$ and $\tilde{\mu}$ a perturbed probability measure. The prior-anchored subspace is defined as follows:
    $$\mathcal{F}_{\text{prior}} = \left\{ g \in \mathcal{F} \ \Big| \ \int_{\mathcal{X}} g(x) \tilde{\mu}(dx) = 0 \right\}$$
    Then, the following statements hold:
    \begin{itemize}
        \item The IPM between the two densities can be evaluated exactly via
        $$\sup_{f \in \mathcal{F}} \left| \int_{\mathcal{X}} f(x) p(dx) - \int_{\mathcal{X}} f(x) q(dx) \right| = \sup_{g \in \mathcal{F}_{\text{prior}}} \left| \int_{\mathcal{X}} g(x) p(dx) - \int_{\mathcal{X}} g(x) q(dx) \right|.$$
        \item Specifically, if we set $\mathcal{X}$ to be the parameter space $\mathcal{X}$ and $p=\tilde{\mu}, q=\tilde{\mu}^y$, the IPM between the prior and posterior can be evaluated exactly over this restricted subspace
        $$\gamma_{\mathcal{F}}(\tilde{\mu}, \tilde{\mu}^y) = \sup_{g \in \mathcal{F}_{\text{prior}}} \left| \int_{\mathcal{X}} g(x) \tilde{\mu}^y(dx) \right| \triangleq D(y)$$
    \end{itemize}
\end{lem}

For the convenience of the theoretical analysis, without loss of generality, we slightly adjust the inducing function class for the 1-Wasserstein distance to $\gF_W^{x_0}:=\{f:f(x_0)=0, \|f\|_L\leq 1\}$. Because shifting a function by a constant does not change its Lipschitz constant, $\gF_W^{x_0}$ still perfectly defines the 1-Wasserstein distance. We denote the corresponding function classes for MMD and energy distance as $\gF_k$ and $\gF_k^{x_0}$, respectively.

To formalize how IPM utilities behave under prior perturbations, the observation noise model must be compatible with the chosen metric's geometry. We assume the noise model is sufficiently smooth and bounded, ensuring the likelihood function interacts well with the function class $\mathcal{F}$.

\begin{assumption}
\label{asmp:compatibility}
    Let \(\mathcal F\) be one of these three classes. For each \(y\in\mathcal Y\), define
\[
e_y(\cdot):=
\begin{cases}
p(y\mid \cdot), & \mathcal F=\mathcal F_k,\\[0.5ex]
p(y\mid \cdot)-p(y\mid x_0), & \mathcal F=\mathcal F_W^{x_0}\ \text{or}\ \mathcal F=\mathcal F_E^{x_0}.
\end{cases}
\]
Assume there exist integrable functions $C_E(y) > 0$ and $C_L(y) > 0$ such that for almost every \(y\in\mathcal Y\) and every \(f\in\mathcal F\),
\[
\frac{1}{C_E(y)}\,e_y\in\mathcal F, \quad
\frac{1}{C_L(y)}\,f(\cdot)\,p(y\mid\cdot)\in\mathcal F.
\]
\end{assumption}

Assumption \ref{asmp:compatibility} imposes a pointwise analytic compatibility condition between the likelihood slice $x \mapsto p(y | x)$ and the function space defining the metric. It asserts that the evidence function belongs to the metric's dual space, and that multiplication by the likelihood is a bounded operator on that space.

\begin{prop}
\label{prop:example}
    Consider the 1D parameter space $\mathcal{X} = [0, R]$ and the additive noise model $Y = X + \eta$, where $\eta \sim \mathcal{N}(0, \sigma^2)$. The likelihood is given by $p(y \mid x) = \phi_\sigma(y-x)$, where $\phi_\sigma(z) = (2\pi\sigma^2)^{-1/2} \exp(-z^2 / 2\sigma^2)$. We use $H^s(\mathcal{X})$ to denote the standard fractional Sobolev space \citep{di2012hitchhiker} and $A_s$ to be the universal constant of a Sobolev multiplier algebra when $s>d/2$ \citep{strichartz1967multipliers}. Using $x_0 = 0$ as the anchor point, Assumption \ref{asmp:compatibility} is satisfied for all three metrics with explicitly integrable local constants:
    \begin{itemize}
        \item For Wasserstein-1 ($W_1$): \[C_E(y) = \sup_{x \in [0,R]} |\phi_\sigma'(y-x)|,\quad C_L(y) = \sup_{x \in [0,R]} \phi_\sigma(y-x) + R \sup_{x \in [0,R]} |\phi_\sigma'(y-x)|.\]
        \item For Mat\'ern-MMD (where $\mathcal{H}_k$ is norm-equivalent to $H^s([0,R])$ with $s > 1/2$, i.e., $c_1 \|\cdot\|_{H^s} \le \|\cdot\|_{\mathcal{H}_k} \le c_2 \|\cdot\|_{H^s}$):
        \[C_E(y) = c_2\|\phi_\sigma(y-\cdot)\|_{H^s([0,R])},\quad C_L(y) = \frac{c_2}{c_1}A_s \|\phi_\sigma(y-\cdot)\|_{H^s([0,R])}.\]
        \item For Energy Distance:
        \[C_E(y) = \|\phi_\sigma'(y-\cdot)\|_{L^2([0,R])},\quad C_L(y) = \sup_{x \in [0,R]} \phi_\sigma(y-x) + \sqrt{R} \|\phi_\sigma'(y-\cdot)\|_{L^2([0,R])}.\]
    \end{itemize}
    Furthermore, for all three metrics, $C_E(y)$ and $C_L(y)$ exhibit Gaussian exponential tail decay as $|y| \to \infty$, guaranteeing that they are integrable over $y$.
\end{prop}
\begin{proof}
    See Appendix \ref{sec:appendix_prior_proof}.
\end{proof}

We emphasize that the example in Proposition \ref{prop:example} can be extended to the high-dimensional case. Consider a compact $d$-dimensional parameter space $\mathcal{X}\subset \mathbb{R}^d$, and the additive noise model $Y=X+\eta$, where $\eta\sim \mathcal{N}(0,\sigma^2 I_d)$. The likelihood is given by $p(y \mid x) = \phi_\sigma(y-x)$, where $\phi_\sigma(z) = (2\pi\sigma^2)^{-d/2} \exp(-z^2 / 2\sigma^2)$. Let $R=\sup_{x\in\mathcal{X}}\|x\|_2$ bound the domain radius. Under the same conditions as in Proposition \ref{prop:example}, we have the following choices. For Wasserstein-1, we have $C_E(y) = \sup_{x\in\mathcal{X}} \|\nabla_x\phi_{\sigma}(y-x)\|_2$ and $C_L(y) = \sup_{x \in \mathcal{X}} \phi_\sigma(y-x) + R \sup_{x \in \mathcal{X}} \|\nabla_x\phi_{\sigma}(y-x)\|_2$. For Mat\'ern-MMD, we have $C_E(y) = c_2 \|\phi_\sigma(y-\cdot)\|_{H^s(\mathcal{X})}$ and $C_L(y) = \frac{c_2}{c_1}A_s\|\phi_\sigma(y-\cdot)\|_{H^s(\mathcal{X})}$. However, the corresponding derivation for the Euclidean Energy Distance is mathematically more delicate, requiring bounds on the posterior multiplier constant in a homogeneous fractional Sobolev space \citep{brasco2021characterisation} by verifying specialized Kato-Ponce inequalities \citep{grafakos2014kato}. While conceptually analogous, we omit this derivation and focus on the Matérn-MMD case, which adequately captures the behavior of RKHS-based metrics in arbitrary dimensions via standard Sobolev multiplier algebras.

We now state our main stability result for prior approximations.

\begin{thm}[Prior Stability Bound] \label{thm:prior}
     Let $\mu$ and $\tilde{\mu}$ be the true and approximate prior distributions, yielding expected utilities $U_{\mathcal{F}}$ and $\tilde{U}_{\mathcal{F}}$, respectively. Under Assumption \ref{asmp:compatibility}, the difference in utility is linearly bounded by the IPM distance between the priors:
    $$|U_{\mathcal{F}} - \tilde{U}_{\mathcal{F}}| \le \bar{K} \gamma_{\mathcal{F}}(\mu, \tilde{\mu}),$$
    where the stability constant is $\bar{K} = 1 + \int_{\mathcal{Y}} \Big( C_L(y) + 2C_E(y)D(y) \Big) dy$.
\end{thm}
\begin{proof}
    See Appendix \ref{sec:appendix_prior_proof}.
\end{proof}

Theorem \ref{thm:prior} links the quality of the prior approximation directly to the resulting utility error. The utility perturbation is bounded linearly by the IPM discrepancy $\gamma_{\mathcal{F}}(\mu, \tilde{\mu})$ between the approximate and true priors. The stability constant $\bar{K}$ captures the sensitivity of this perturbation and depends primarily on the posterior discrepancy $D(y)$. This reveals that the penalty incurred by approximating the prior depends strictly on how severely the observation $y$ shifts the posterior distributions across the chosen test functions.

In addition, this formulation highlights a key difference between bounded and unbounded metrics. For bounded IPMs , the test functions are capped, meaning the posterior discrepancy $D(y)$ is universally bounded. As a result, the stability constant $\bar{K}$ remains finite without additional restrictive assumptions on the true prior's tail decay or the forward model's growth rate. In contrast, for unbounded IPMs, $D(y)$ can grow arbitrarily large if the forward model causes severe spatial warping. Ensuring a finite $\bar{K}$ in these cases typically requires stronger conditions, such as assuming a sub-Gaussian prior and a globally Lipschitz forward model to restrict this growth. We provide a deeper investigation into the mechanics of $D(y)$ and its practical implications in Section \ref{sec:understanding}.

\paragraph{Contrast with KL Divergence:} Standard EIG, which relies on KL divergence, is fundamentally incompatible with the discrete prior approximations commonly used in modern computational pipelines. Because KL divergence requires the distribution to be absolutely continuous with respect to one another, the divergence between a continuous prior and a discrete empirical measure is technically undefined. In contrast, because IPMs operate directly on probability measures via spatial transport or kernel embeddings rather than density ratios, they provide a much more natural theoretical justification for the empirical particle approximations used in practice.

\subsection{Deeper Understanding of $\omega_{\gF}(x)$ and $D(y)$}
\label{sec:understanding}
Having established the main stability bounds, we now examine the two geometric quantities that govern them: the test-function growth profile $\omega_{\mathcal{F}}(x)$ and the posterior discrepancy $D(y)$. In this section, we compare their behavior across a hierarchy of IPMs: bounded metrics (MMD, for which $\omega_{\mathcal{F}}(x)=\mathcal{O}(1)$), sublinear metrics (energy distance, for which $\omega_{\mathcal{F}}(x)=\mathcal{O}(\|x\|^{1/2})$), and linear metrics (Wasserstein-1, for which $\omega_{\mathcal{F}}(x)=\mathcal{O}(\|x\|)$). We then show that controlling these quantities naturally leads to the assumptions used in related work, such as globally Lipschitz forward models and sub-Gaussian priors \citep{helin2025bayesian}. A mathematically rigorous discussion is deferred to Appendix \ref{sec:math_understanding}.

\paragraph{Likelihood Stability} In the likelihood-stability bound, sensitivity to perturbations in the data misfit is governed by the test-function growth profile $\omega_{\mathcal{F}}(x)$. Assume a Gaussian observation model $y = G(x) + \eta$ and a relative modeling perturbation satisfying $\|\delta(x)\| \le \epsilon(1 + \|G(x)\|)$. A first-order Gaussian shift estimate shows that the likelihood perturbation enters the utility bound through the weighted prior integral $
\int_{\mathcal X} \omega_{\mathcal F}(x)\bigl(1+\|G(x)\|\bigr)\,\mu(dx).$
If $G$ is globally Lipschitz, then $\|G(x)\|\lesssim 1+\|x\|$, so the required prior integrability is determined by the growth of $\omega_{\mathcal F}(x)$. For bounded-kernel MMD, this yields a sufficient condition of a finite first moment. For the anchored energy-distance class, the corresponding sufficient condition is a finite $3/2$-moment. For Wasserstein-1, the same argument gives a finite second moment. This makes the basic hierarchy precise: as the IPM becomes less bounded in space, the likelihood-stability bound requires stronger prior integrability.

\noindent 

\paragraph{Prior Stability} The prior-stability term behaves differently. The key quantity is $
D(y)=\gamma_{\mathcal F}(\mu^y,\tilde\mu^y)$,
which depends on the posterior reweighting induced by the likelihood. For bounded-kernel MMD, the test functions are uniformly bounded in the supremum norm, so $D(y)$ is uniformly bounded and no tail-growth analysis is required. By contrast, unbounded IPMs involve transporting mass, and the local sensitivity is determined by the gradient of the effective integrand $h_y(x) = f(x)\exp(-\Phi(x;y))$.
Under a differentiable globally Lipschitz forward model, the gradient of the potential grows linearly, that is, as $\mathcal{O}(\|x\|)$. Consequently, the product rule introduces an additional factor of $\|\nabla_x\Phi(x;y)\|$ into the effective test class. For Wasserstein-1, the envelope grows as $\mathcal{O}(\|x\|)$, so the resulting gradient grows quadratically, leading to a term of order $\mathcal{O}(\|x\|^2)$. For the anchored energy-distance class, the envelope grows as $\mathcal{O}(\|x\|^{1/2})$, so the effective gradient grows superlinearly, namely as $\mathcal{O}(\|x\|^{1.5})$, depending on the precise formulation of the class. This leads to the same qualitative ordering as above: bounded-kernel MMD is the least sensitive to prior tails, the anchored energy-distance class is intermediate, and Wasserstein-1 is the most tail-sensitive. In particular, strong tail assumptions, such as sub-Gaussian priors, arise naturally as convenient sufficient conditions in Wasserstein-based analyses.

\section{Experiments}
\label{sec:experiments}
In this section, we empirically evaluate our IPM-based BOED core framework and then its beyond-IPM plug-and-play extensions in high-dimensional settings. Building on benchmark tasks of \cite{foster2019variational}, we consider two settings of increasing complexity: (i) analytically tractable A/B testing problem, and (ii) preference learning model. In addition, to further illustrate the plug-and-play nature of our framework, we consider (iii) high-dimensional linear-Gaussian benchmark and (iv) high-dimensional sign-ambiguous benchmark.

\subsection{Analytical Validation: A/B Testing}
\label{sec:exp_ab_test}

We begin with an A/B testing scenario in which the expected utility can be computed exactly, allowing a rigorous validation of our formulation. Specifically, we consider two alternatives, denoted A and B. The design variable $\xi = (n_A, n_B)$ allocates a total budget of $N = n_A + n_B$ samples between the two groups. The parameters $x_A$ and $x_B$ denote the mean responses of the respective groups. The quantity of interest is the treatment effect, $\delta := x_B - x_A,$
which measures the improvement of variant B over A and is invariant under global shifts. Our goal is therefore to identify a design $\xi$ that maximally reduces uncertainty about $\delta$.

We assume a Gaussian prior $x \sim \mathcal{N}(0, \Sigma_0)$ and an observation model $y | x, \xi \sim \mathcal{N}(X_\xi x, I)$, where $X_\xi \in \mathbb{R}^{N \times 2}$ is the design matrix. The first $n_A$ rows of $X_\xi$ are $(1, 0)$ and the remaining $n_B$ rows are $(0, 1)$. By Bayes' theorem, the posterior is given by $x | y, \xi \sim \mathcal{N}(\mu_1, \Sigma_1)$, where $\Sigma_1 = (\Sigma_0^{-1} + X_\xi^\top X_\xi)^{-1}$ and $\mu_1 = \Sigma_1 X_\xi^\top y$.

For this experiment, we set the prior covariance to $\Sigma_0 = \mathrm{diag}(\sigma_A^2, \sigma_B^2)$ with $\sigma_A = 5$ and $\sigma_B = 1.8$. Consequently, the prior over the treatment effect is $\delta \sim \mathcal{N}(0, \sigma_A^2 + \sigma_B^2)$. The posterior parameters then simplify to:
$$
\Sigma_1 = \begin{pmatrix} \frac{1}{n_A + 1/\sigma_A^2} & 0 \\ 0 & \frac{1}{n_B + 1/\sigma_B^2} \end{pmatrix}, \quad 
\mu_1 = \begin{pmatrix} \frac{\sum_{i \in A} y_i}{n_A + 1/\sigma_A^2} \\ \frac{\sum_{i \in B} y_i}{n_B + 1/\sigma_B^2} \end{pmatrix}.
$$
Denoting the marginal posterior distributions by $x_A \mid y, \xi \sim \mathcal{N}(m_A, v_A)$ and $x_B \mid y, \xi \sim \mathcal{N}(m_B, v_B)$, we obtain $\delta \mid y, \xi \sim \mathcal{N}(m_B - m_A,\, v_A + v_B)$.
The $W_1$ distance between the prior and posterior distributions of $\delta$ is
\[
W_1\!\left(\mathcal{N}(0, \sigma_A^2 + \sigma_B^2), \mathcal{N}(m_B - m_A, v_A + v_B)\right)
=
\mathbb{E}\left|\, m_A - m_B + \left(\sqrt{\sigma_A^2 + \sigma_B^2} - \sqrt{v_A + v_B}\right) Z \right|,
\]
where $Z \sim \mathcal{N}(0,1)$. Since $m_A - m_B$ is a zero-mean Gaussian random variable, and the law of total variance gives $\mathrm{Var}(\delta)
=
\mathrm{Var}(\mathbb{E}[\delta \mid y]) + \mathbb{E}[\mathrm{Var}(\delta \mid y)]$,
we obtain
\[
\mathrm{Var}(m_B - m_A) = \sigma_A^2 + \sigma_B^2 - (v_A + v_B).
\]
Therefore, the quantity inside the absolute value is itself a centered Gaussian random variable, which yields the closed-form expected utility
\[
U(\xi)
=
\sqrt{\frac{2}{\pi}}
\sqrt{
2(\sigma_A^2 + \sigma_B^2)
-
2\sqrt{\sigma_A^2 + \sigma_B^2}\sqrt{v_A + v_B}
}.
\]
Since the prior variances $\sigma_A^2$ and $\sigma_B^2$ are constants, maximizing $U(\xi)$ is equivalent to minimizing the posterior variance
\[
v_A + v_B
=
\frac{1}{n_A + 1/\sigma_A^2}
+
\frac{1}{n_B + 1/\sigma_B^2}.
\]
For $N=10$, the continuous relaxation yields an optimal allocation of $n_A^\ast \approx 5.13$. The optimal discrete design is therefore $n_A = 5$ and $n_B = 5$.

For comparison, we also evaluate the classical EIG based on KL divergence. By a similar derivation, the KL utility reduces to
\[
U_{\mathrm{KL}}(\xi)
=
\frac{1}{2}\log\frac{\sigma_A^2 + \sigma_B^2}{v_A + v_B}.
\]

Thus, maximizing the KL utility is also equivalent to minimizing $v_A + v_B$. Both criteria therefore identify the same optimal design, namely $(n_A, n_B) = (5,5)$. This confirms that the Wasserstein utility preserves the correct global optimum in this standard setting while benefiting from the improved stability properties established in Section \ref{sec:stability_analysis}.

We use this example mainly as a sanity check: it shows that replacing KL by an IPM does not distort the global optimum in a standard setting where the utility can be analyzed exactly. Differences in the shape of the resulting utility landscapes are explored empirically in the next subsection.

\subsection{Computational Performance: Preference Learning}
To evaluate our framework in a setting without a closed-form posterior and with nonlinear observations, we next consider a preference learning model. In this more challenging setting, we compare IPM-based utilities with the classical KL-based criterion from the perspectives of both computational efficiency and optimization reliability.

\paragraph{Model Formulation.} We introduce a latent parameter $x \in \mathbb{R}$ representing a user's intrinsic preference threshold, together with a continuous scalar design variable $\xi \in [\xi_{\min}, \xi_{\max}]$ representing a characteristic of the proposed item. We place a Gaussian prior $x \sim \mathcal{N}(\mu_x, \sigma_x^2)$,
with default values $\mu_x = -20$ and $\sigma_x = 20$. Conditional on the parameter-design pair $(x,\xi)$, we model the latent utility $\eta$ as
\[
\eta \mid x, \xi \sim \mathcal{N}(\xi - x, \sigma_\eta^2(1 + |\xi|)^2),
\]
where $\sigma_\eta = 1$. The noise variance therefore grows quadratically with the magnitude of the design, which naturally penalizes extreme design choices. The final observed response is obtained by passing the latent utility through a sigmoid link and then truncating the result for numerical stability:
\[
y = \max(\epsilon, \min(1 - \epsilon, \mathrm{sigmoid}(\eta))),
\]
with $\epsilon = 0.01$. This yields a bounded continuous observation $y \in [\epsilon, 1-\epsilon]$, which can be interpreted as a noisy preference probability.

\paragraph{Experimental Setup and Computational Speedup.} We discretize the design space on a uniform grid from $-80$ to $80$ with step size $2$, yielding $81$ candidate designs.

\begin{figure}[htb!]
    \centering
    \includegraphics[width=0.48\linewidth]{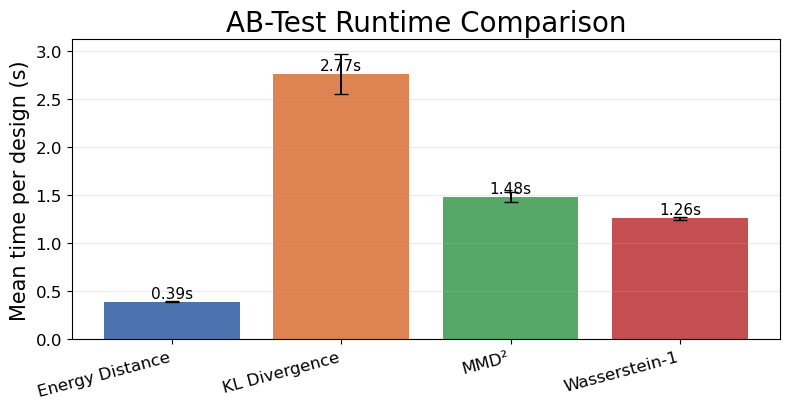}
    \hfill
    \includegraphics[width=0.48\linewidth]{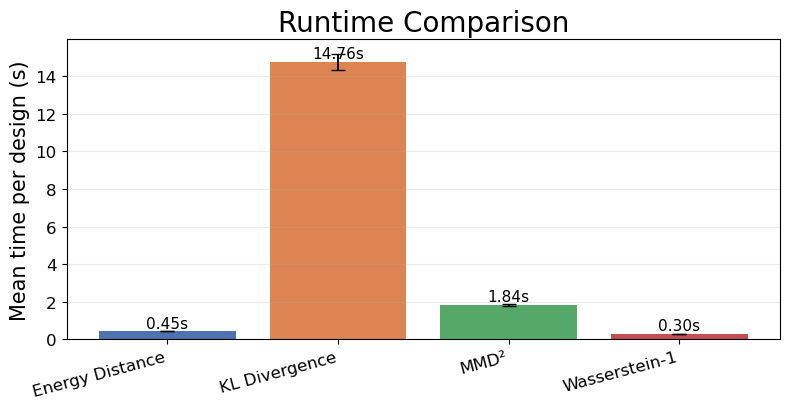}
    
    \caption{Computational runtime comparison between KL-based EIG and IPM-based utilities. The plots report the per-design evaluation time for both the A/B testing model (left) and the preference learning model (right), highlighting the substantial speedup achieved by density-free IPMs.}  \label{fig:preference_runtime}
\end{figure}

As shown in Figure \ref{fig:preference_runtime}, the empirical results strongly support the computational advantage of the density-free IPM approach. Under the current implementation for this experiment, the KL-based baseline is substantially slower because it requires an additional density-estimation step before evaluating the log-ratio utility, together with bandwidth tuning that is both computationally expensive and numerically sensitive. By contrast, the IPM utilities considered here are evaluated directly from samples through pairwise distances or kernels, which avoids this intermediate density-estimation stage. As a result, in this experiment the W1-based utility requires only 0.3 seconds per design, whereas the KL baseline requires about 14 seconds per design. Figure \ref{fig:preference_runtime} also includes the runtime for the A/B testing experiment in Section \ref{sec:exp_ab_test}, showing that IPMs retain a consistent efficiency advantage in both analytically tractable and numerically approximated settings.

\paragraph{Utility Landscape and Optimization Reliability.}

\begin{figure}[htb!]
    \centering
    \includegraphics[width=\linewidth]{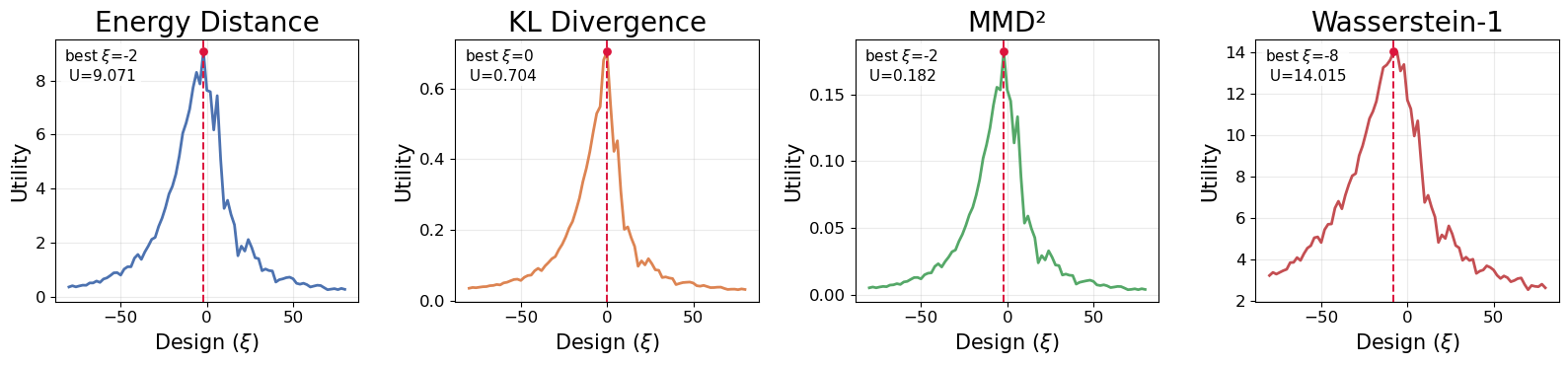}
    \caption{Expected utility landscapes across candidate designs for preference learning.}
    \label{fig:preference_utility}
\end{figure}

Beyond computational speed, a central practical observation is that IPMs also improve the geometry of the optimization landscape. Figure \ref{fig:preference_utility} displays the expected utility curves across the candidate designs and shows how the choice of utility metric changes the structure of the design objective.

\label{sec:preference}

To formalize this optimization reliability, we define the \textit{high-utility design region}
\[
R_t = \{ d : U(d) \ge t \cdot \max U \}.
\]
Conceptually, this is analogous to a Highest Posterior Density (HPD) credible region in Bayesian inference \citep{lee1989bayesian}, but applied to the experimental design space: it characterizes the basin of near-optimal and highly informative designs. The size of this region is measured by the number of discrete design points contained in $R_t$, while the concentration score $S_R$ quantifies how localized the utility mass is over the design space.

\begin{figure}[htb!]
    \centering
    \includegraphics[width=0.9\linewidth]{figs/pref_good_.8.png}
    
    \vspace{1em}
    \begin{tabular}{l | c | c | l}
        \textbf{Metric} & $\boldsymbol{S_R}$ & \textbf{Region $\boldsymbol{R_{0.80}}$} & \textbf{Details} \\
        \hline
        KL Divergence & $0.1213$ & $\subseteq [-2, 0]$ & (2 pts $\ge 80\%$ of max) \\
        Energy Distance & $0.3186$ & $\subseteq [-8, 6]$ & (7 pts $\ge 80\%$ of max)
        \\
        MMD$^2$ & $0.2158$ & $\subseteq [-6, 0]$ & (4 pts $\ge 80\%$ of max) \\
        Wasserstein-1 & $0.2877$ & $\subseteq [-18, 2]$ & (\textbf{11 pts} $\ge 80\%$ of max) \\
    \end{tabular}
    \vspace{0.5em}
    \caption{Characterization of the high-utility design region $R_{0.80} = \{d : U(d) \ge 0.8 \cdot \max U\}$. At the threshold $0.8\,\max U$, IPM-based utilities induce a substantially broader and more robust basin of near-optimal designs than the classical KL divergence.}

    \label{fig:0.8goodregioin}
\end{figure}

\begin{figure}[htb!]
    \centering
    \includegraphics[width=\linewidth]{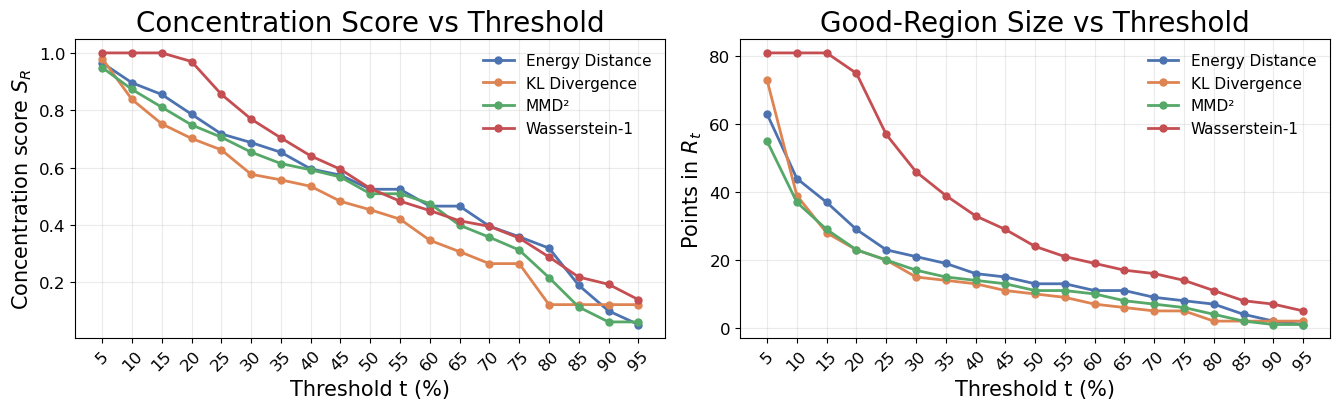}
    \caption{Sensitivity analysis of the high-utility design region across threshold levels $t$. For a broad range of practically relevant thresholds, IPM-based utilities retain a larger number of acceptable design points (right) together with a smoother concentration profile (left), indicating greater optimization stability than KL divergence.}

    \label{fig:pref_good_region}
\end{figure}

Figures \ref{fig:0.8goodregioin} and \ref{fig:pref_good_region} show that KL- and IPM-based utilities differ not only in value, but also in landscape geometry. At the representative threshold $t=0.8$, IPM-based utilities retain a substantially larger superlevel set $\{d : U(d) \ge t\,U_{\max}\}$, indicating that highly informative designs occupy a much broader portion of the design space. By contrast, the KL-based criterion concentrates acceptable designs in a much narrower region around the maximizer. This distinction is important for downstream optimization. In practice, the selected design is affected by grid discretization, Monte Carlo error, approximate posterior computation, and imperfect search. A broader near-optimal basin therefore makes the design recommendation less sensitive to such perturbations, because many nearby designs still achieve utility close to the optimum. In this sense, IPM-based objectives transform the problem from finding a single sharp but unstable peak to identifying a much more stable set of highly competitive designs.

We now analyze whether this geometric property translates into a practical
advantage when the design optimization is approximate. To test whether a wider near-optimal region leads to more robust design
selection under approximate optimization, we evaluate each metric under
\emph{coarse-grid selection}. We restrict the candidate set to every $k$-th
point of $\Xi$ for $k \in \{2, 4, 6, 8\}$. Denote the reference utility as $U_{\mathrm{ref}}$ and the optimal design as $\xi^*$.
For a given $k$ there are $k$ distinct coarse grids, obtained by starting
at offset $s \in \{0, 1, \ldots, k-1\}$.
For each offset we select the design $\hat\xi$ with the highest reference utility on that coarse set, and measure the
normalized regret
\begin{equation*}
  \bar{r}(\hat\xi) \;=\;
  \frac{U_{\mathrm{ref}}(\xi^*) - U_{\mathrm{ref}}(\hat\xi)}
       {U_{\mathrm{ref}}(\xi^*) - \min_{\xi}U_{\mathrm{ref}}(\xi)}.
  \label{eq:norm_regret}
\end{equation*}
We report the mean and maximum of $\bar{r}$ over all $k$ offsets.
Because selection is based on the reference utility itself, this protocol isolates the effect of grid coarsening alone, without additional Monte Carlo noise.
Averaging over offsets removes any dependence on a particular grid
alignment and yields a more general measure of robustness.

\begin{table}[h]
  \centering
  \caption{Mean and maximum normalised regret $\bar{r}(\hat\xi)$ averaged over all $k$ offsets. Lower regret indicates better recovery under coarse-grid restriction. $|R_{0.80}|$ stands for the number of points within the high-utility design region as shown in Figure \ref{fig:0.8goodregioin}.}
  \label{tab:coarse_design}
  \begin{tabular}{lc cccc}
    \toprule
    & & \multicolumn{3}{c}{Mean norm.\ regret $\bar{r}$ (max in parentheses)} \\
    \cmidrule(lr){3-6}
    Metric & $|R_{0.80}|$
           & $k=2$ & $k=4$ & $k=6$ & $k=8$ \\
    \midrule
    KL divergence
      & 2
      & 0.018 (0.035) & 0.121 (0.231) & 0.180 (0.337) & 0.234 (0.419) \\
    Energy distance
      & 7
      & 0.068 (0.135) & 0.093 (0.151) & 0.117 (0.168) & 0.148 (0.299) \\
    MMD$^2$
      & 4
      & 0.080 (0.160) & 0.118 (0.161) & 0.150 (0.225) & 0.195 (0.383) \\
    Wasserstein-1
      & \textbf{11}
      & \textbf{0.001} (0.002) & \textbf{0.023} (0.056)
      & \textbf{0.038} (0.081) & \textbf{0.054} (0.134) \\
    \bottomrule
  \end{tabular}
\end{table}

Table \ref{tab:coarse_design} reports the resulting mean and maximum normalized regret across offsets. The results provide systematic, alignment-robust evidence in this preference-learning example that broader high-utility regions are associated with smaller regret under coarse-grid selection. Across all reported coarsening levels, Wasserstein-1 achieves the lowest mean and worst-case regret, while KL exhibits the strongest degradation as the grid becomes coarser. Energy Distance and MMD$^2$ lie in between. Moreover, for each $k$, both the mean and maximum regret are ordered consistently with the size of the $0.80$-superlevel set. This indicates that the width of the near-optimal design region captures an operationally meaningful aspect of robustness to design discretization.

The coarse-grid study above provides concrete one-step evidence that broader high-utility regions can reduce the sensitivity of design selection to discretization. This observation may also be relevant for sequential OED, where design choices are repeatedly made under approximate posterior updates and imperfect numerical optimization. A broader near-optimal region may then help reduce the propagation of local optimization errors across stages. We emphasize, however, that this sequential interpretation remains suggestive in the present work, and a direct empirical study of multi-stage robustness is left for future work.

\subsection{High-Dimensional Plug-and-Play Extensions Beyond IPMs: Scalability and Robustness}
\label{sec:highd_exp}
The experiments in previous sections show that IPM-based utilities are numerically stable, computationally efficient, and more reliable than classical KL-based criteria. We now show that the same sample-based BOED pipeline can also accommodate geometry-aware discrepancies beyond the IPM class. First, we study whether the sample-based framework can be combined directly with scalable neural estimators in a high-dimensional problem where exact ground truth is available. Second, we test whether this approach remains reliable when the posterior geometry becomes strongly non-Gaussian. To this end, we consider two high-dimensional benchmarks with complementary structure: a $64$-dimensional linear-Gaussian model with a tractable unimodal posterior, and a $32$-dimensional sign-ambiguous model with a bimodal posterior. Together, these experiments show that the proposed framework is not only scalable, but also robust to posterior features that are known to cause difficulty for KL-based approximations: density estimation fails under the curse of dimensionality \citep{li2024expected}, and standard nested Monte Carlo (NMC) estimators can suffer from exponentially growing variance \citep{foster2019variational}.

\paragraph{Plug-and-Play Neural Estimators.} To evaluate the design utility empirically (simulating a scenario where the closed-form equation is unknown), we utilize the OT-ICNN method in \citet{makkuva2020optimal}. This approach parameterizes the Kantorovich dual potentials using Input Convex Neural Networks (ICNNs), which enforce the convexity structure required by optimal transport. Importantly, incorporating this neural estimator requires no modification of the BOED framework itself. For each design $\xi$ and sampled observation $y$, we draw fresh minibatches at every optimization step. We then train two ICNN potentials under the semidual objective, alternating multiple updates of the transport network for each update of the critic, together with a convexity regularization penalty on the constrained weights. In these two experiments, we use three hidden layers of width $256$ for benchmark I and $128$ for benchmark II, a batch size of $128$, the Adam optimizer with learning rate $10^{-4}$, and a convexity penalty coefficient of $0.1$. Final utility estimates are obtained by averaging the learned OT objective over several fresh evaluation minibatches. For comparison, we estimate the KL-based EIG using nested Monte Carlo with $64$ outer samples and $512$ inner prior samples per outer sample for benchmark I and Gaussian variational methods with $50000$ samples in total for benchmark II.


\paragraph{Benchmark I: A High-Dimensional Linear-Gaussian Model.} We first consider a linear-Gaussian design problem in dimension $p = 64$. The prior is $x \sim \mathcal{N}(0, I_p)$, and the design variable is a scalar gain $\xi \in \Xi = \{0.5, 2.0, 6.0\}$. Given a design $\xi$, the observation model is
\[
y \mid x, \xi \sim \mathcal{N}(\xi x, \sigma^2 I_p),
\]
with $\sigma = 0.1$. Because this model is conjugate \citep{raiffa2000applied}, the posterior is available in closed form:
\[
p(x \mid y, \xi) = \mathcal{N}(\mu_{\xi}(y), \Sigma_{\xi}),
\]
where
\[
\mu_{\xi}(y) = \frac{\xi}{\xi^2 + \sigma^2} y,
\qquad
\Sigma_{\xi} = \frac{\sigma^2}{\xi^2 + \sigma^2} I_p.
\]

This benchmark serves as a clean test of scalability. Since the posterior is unimodal and analytically tractable, it isolates the effect of dimension without introducing additional approximation error from complicated posterior geometry. It also allows us to compute the exact expected $2$-Wasserstein utility in closed form:
\[
U(\xi) = \mathbb{E}_{y \mid \xi}\!\left[W_2^2\bigl(\mathcal{N}(0, I_p), \mathcal{N}(\mu_{\xi}(y), \Sigma_{\xi})\bigr)\right]
= 2p \left(1 - \frac{\sigma}{\sqrt{\xi^2 + \sigma^2}}\right),
\]
which provides a ground-truth reference for evaluating the neural estimator.

\paragraph{Results on Benchmark I.}
We compare the OT-ICNN Wasserstein estimator with a standard prior-sampling nested Monte Carlo estimator for the classical EIG. Table \ref{table:toy_boed_icnn} reports the results for the candidate designs.

\begin{table}[htb!]
\centering
\caption{Comparison with exact utilities in the $64$-dimensional linear-Gaussian BOED problem. ICNN is evaluated against the exact Wasserstein utility, while nested Monte Carlo is evaluated against the exact expected information gain (EIG). The relative error is defined by $|\hat{U}-U^\star|/|U^\star|$.}
\begin{tabular}{lccc|ccc}
\toprule
 & \multicolumn{3}{c|}{$W_2^2$ utility} & \multicolumn{3}{c}{EIG utility} \\
Design & Exact & ICNN & Rel.\ err. & Exact & Nested MC & Rel.\ err. \\
\midrule
$\xi=0.5$ & $1.03 \times 10^2$ & $1.01 \times 10^2$ & \bm{$2.18 \times 10^{-2}$} & $1.04 \times 10^2$ & $9.49 \times 10^2$ & $8.10 \times 10^0$ \\
$\xi=2$   & $1.22 \times 10^2$ & $1.21 \times 10^2$ & \bm{$2.70 \times 10^{-3}$} & $1.92 \times 10^2$ & $1.54 \times 10^4$ & $7.91 \times 10^1$ \\
$\xi=6$   & $1.26 \times 10^2$ & $1.25 \times 10^2$ & \bm{$3.10 \times 10^{-3}$} & $2.62 \times 10^2$ & $1.34 \times 10^5$ & $5.12 \times 10^2$ \\
\bottomrule
\end{tabular}
\label{table:toy_boed_icnn}
\end{table}

The results show a clear contrast between the two approaches. The OT-ICNN estimator scales effectively to $p=64$, closely matches the exact expected $W_2^2$ utility, and recovers the correct ranking of the candidate designs, with relative errors of $0.0218$, $0.0027$, and $0.0031$. By contrast, the naive nested Monte Carlo estimator for the KL-based EIG becomes highly inaccurate in this high-dimensional, low-noise regime, with relative errors ranging from $8.10$ to $5.12 \times 10^2$. Thus, even in a mild unimodal problem, standard EIG estimation can become unreliable in high dimension unless substantial problem-specific variance-reduction techniques are introduced. In contrast, the sample-based BOED formulation allows scalable neural estimators to be integrated directly into the framework, making high-dimensional utility estimation feasible without altering the underlying method.

\paragraph{Benchmark II: A High-Dimensional Sign-Ambiguous Bimodal Model.}
We now turn to a second high-dimensional benchmark that stresses a different issue. The challenge here is not conjugate high-dimensional scaling, but severe posterior non-Gaussianity. Specifically, we construct a $32$-dimensional sign-ambiguous model in which the posterior is bimodal and therefore poorly matched by Gaussian variational approximations.

Let $x \in \R^{32}$. Each coordinate has an independent bimodal prior
\[
p(x_i) = \frac{1}{2}\mathcal{N}(-\mu_i,\tau^2)
       + \frac{1}{2}\mathcal{N}(\mu_i,\tau^2),
\]
with $\tau = 0.35$. A design $\xi \subseteq [32]$, with $|\xi|=8$, observes the selected coordinates through
\[
y_i = x_i^2 + \epsilon_i,
\quad \epsilon_i\sim \mathcal{N}(0,\sigma^2), \quad
\sigma = 0.20.
\]

Because the map $x_i \mapsto x_i^2$ removes the sign of $x_i$, the likelihood is insensitive to the sign of $x_i$. As a result, the posterior is sign-ambiguous, and the degree of non-Gaussianity increases with $\mu_i$. The prior and likelihood both factorize across coordinates, so the exact design-level utilities decompose as sums of one-dimensional contributions. Table~\ref{tab:coord-types} reports the resulting per-coordinate reference values for four coordinate types.

\begin{table}[h]
  \centering
  \caption{%
    Per-coordinate 1-D reference utilities for each $\mu_i$ type.
    Design-level utilities are sums over the eight observed coordinates.
  }
  \label{tab:coord-types}
  \smallskip
  \begin{tabular}{llccc}
    \toprule
    Type   & Coord.\ indices & $\mu_i$ &
    $\mathrm{EIG}_{1d}$ & $W_{2,1d}^2$ \\
    \midrule
    Strong & 1--8   & 4.0 & 2.637 & 0.243 \\
    Medium & 9--16  & 2.0 & 1.941 & 0.200 \\
    Weak   & 17--24 & 1.0 & 1.251 & 0.167 \\
    Null   & 25--32 & 0.1 & 0.265 & 0.031 \\
    \bottomrule
  \end{tabular}
\end{table}
We consider five candidate designs, each observing $8$ coordinates, chosen to span a broad range of information profiles:
\textbf{A}~(8 strong), \textbf{B}~(8 medium), \textbf{C}~(8 weak), \textbf{D}~(4 strong $+$ 4 null), and \textbf{E}~(8 null).

\begin{table}[h]
  \centering
  \caption{%
    Utilities and design rankings on the sign-ambiguous benchmark
    ($d=32$, $|\xi|=8$, seed 42).
    Rankings in parentheses ($1=$ best).
  }
  \label{tab:highd-results}
  \smallskip
  \setlength{\tabcolsep}{5pt}
  \begin{tabular}{llrrrr}
    \toprule
    & &
    \multicolumn{2}{c}{EIG} &
    \multicolumn{2}{c}{$\mathcal{W}_2^2$} \\
    \cmidrule(lr){3-4}\cmidrule(lr){5-6}
    Design & Composition
      & Exact & Gaussian VB
      & Exact & OT-ICNN \\
    \midrule
    A & 8 strong
      & $21.09\ (1)$ & $-13.91\ (5)$
      & $1.943\ (1)$ & \bm{$1.608\ (1)$} \\
    B & 8 medium
      & $15.53\ (2)$ & $\ -8.27\ (4)$
      & $1.596\ (2)$ & \bm{$1.585\ (2)$} \\
    C & 8 weak
      & $10.00\ (4)$ & $\ -2.53\ (2)$
      & $1.337\ (3)$ & \bm{$1.249\ (3)$} \\
    D & 4 strong $+$ 4 null
      & $11.61\ (3)$ & $\ -6.34\ (3)$
      & $1.097\ (4)$ & \bm{$0.793\ (4)$} \\
    E & 8 null
      & $\ \ 2.12\ (5)$ & $\ +1.23\ (1)$
      & $0.251\ (5)$ & \bm{$0.161\ (5)$} \\
    \midrule
    \multicolumn{2}{l}{\textit{Ranking}} &
      ABDCE &
      ECDBA &
      ABCDE &
      \textbf{ABCDE} \\
    \bottomrule
  \end{tabular}
\end{table}

\paragraph{Results on Benchmark II.}
To evaluate KL-based design under posterior misspecification, we use the Barber--Agakov lower bound \citep{barber2004algorithm},
\begin{equation*}
  \mathrm{EIG} (\xi) \;\ge\;
  H[p(x)] + \E_{p(x,y)}\left[\log q_\phi(x \mid y)\right],
  \label{eq:ba-bound}
\end{equation*}
where $q_\phi$ is a Gaussian variational approximation. This lower bound is accurate only when $q_\phi$ approximates the true posterior well. In the present model, however, the posterior is bimodal, so any single Gaussian must place mass between the two modes rather than on them. A straightforward calculation can show that for large $\mu_i$ the resulting per-coordinate lower bound is negative. This explains the numerical results in Table \ref{tab:highd-results} that the bound becomes most negative on the most informative coordinates, while remaining close to zero on the nearly uninformative null coordinates. This completely reverses the design ranking.

By contrast, the OT-ICNN estimator within our Wasserstein BOED framework remains stable. Using $50$ Monte Carlo samples, it produces the ranking
\[
\text{A}>\text{B}>\text{C}>\text{D}>\text{E},
\]
which matches the true $W_2^2$ ordering exactly. The relatively larger errors on designs A and D reflect the more difficult geometry of well-separated or mixed-type posteriors, but the separation between adjacent designs is large enough for all designs to be ranked correctly.

This benchmark also highlights a conceptual difference between EIG and $W_2^2$. Exact EIG ranks D above C ($11.61$ versus $10.00$) because the four null coordinates in design D still contribute $1.06$. Exact $W_2^2$, however, ranks C above D because the same null coordinates contribute only $0.12$, which is less than $3\%$ of design C's total utility. Thus, $W_2^2$ penalizes the allocation of design budget to coordinates that induce only negligible posterior change more strongly than EIG.

\section{Conclusion}
\label{sec:conclusion}
In this work, we have revisited Bayesian optimal experimental design through the lens of utility design. Our starting point is a structural limitation of the classical expected information gain (EIG) criterion: as a KL-based utility, it depends on a log-density ratio and can therefore be sensitive to rare events, tail underestimation, and model approximation errors. To address this issue, we have introduced a geometry-aware BOED framework based on Integral Probability Metrics (IPMs), with particular emphasis on the 1-Wasserstein distance, Maximum Mean Discrepancy, and Energy Distance.

Our theoretical analysis has established systematic stability guarantees and has shown that IPM-based utilities naturally mitigate the tail sensitivity inherent in density-ratio-based metrics. In particular, we have shown that IPMs admit stability bounds controlled directly by the error in the surrogate likelihood or the prior distribution. Empirically, we have demonstrated that the proposed framework yields more stable and reliable behavior than classical EIG, measured via broader near-optimal design regions, smoother utility landscapes, and improved robustness. By incorporating a plug-and-play neural optimal transport estimator, we have also extended the same sample-based BOED template to high-dimensional design spaces where commonly used EIG estimators become unreliable.

\paragraph{Future Work} The modular and plug-and-play structure of our IPM-based framework opens several directions for future research. A natural next step is to extend the present static BOED framework to sequential OED and active learning settings \citep{he2008active,dror2008sequential,mackay1992information}. Our coarse-grid robustness results suggest that utility-landscape geometry may play an important role when design choices must be made repeatedly under approximate optimization, but a direct empirical study of sequential robustness remains to be carried out. Moreover, because our framework relies on sample-based discrepancy evaluation rather than density tracking, it is naturally compatible with modern generative modeling techniques. Incorporating advanced sampling methods, such as diffusion models \citep{ho2020denoising} and flow matching \citep{lipman2022flow}, into the utility estimators may further improve scalability and fidelity for highly complex and multimodal target distributions.

More broadly, the modular structure of the proposed framework may also be relevant beyond classical BOED. For example, related ideas may be useful for data selection in large-scale model training and for active preference learning in human-in-the-loop systems \citep{sorscher2022beyond,xie2023data,rafailov2023direct,muldrew2024active}. We hope that the present work provides a useful step toward more robust and scalable design criteria in resource-constrained and high-dimensional inference problems.



\acks{HY was partially supported by the US National Science Foundation under awards IIS-2520978, GEO/RISE-5239902, the Office of Naval Research Award N00014-23-1-2007, DOE (ASCR) Award DE-SC0026052, and
the DARPA D24AP00325-00. DW thanks Chenguang Duan for the discussion of sampling. Approved for public release; distribution is unlimited.}



\appendix



\section{Proof of Likelihood Stability}
\label{sec:appendix_likelihood_proof}
\begin{proof}[Proof of Theorem \ref{thm:likelihood}]
    By definition, the expected utility can be written as:
    $$U_{\mathcal{F}} = \int_{\mathcal{Y}} \sup_{f \in \mathcal{F}} \left| \int f(x) p(y|x) \mu(dx) - p(y) \int f(x) \mu(dx) \right| dy$$
    Define the unperturbed and perturbed inner operators for a given $y$ and $f$:
    $$\begin{aligned} A(f, y) &= \int_{\mathcal{X}} f(x)p(y|x)\mu(dx) - p(y) \int_{\mathcal{X}} f(x)\mu(dx), \\ A^*(f, y) &= \int_{\mathcal{X}} f(x)p_*(y|x)\mu(dx) - p_*(y) \int_{\mathcal{X}} f(x)\mu(dx). \end{aligned}$$
    Then, the difference in utility can be bounded by $$|U_{\mathcal{F}} - U_{\mathcal{F}}^*| \le \int_{\mathcal{Y}} \left| \sup_{f} |A(f,y)| - \sup_{f} |A^*(f,y)| \right| dy.$$
    Using the reverse triangle inequality for suprema, we have:
    $$|U_{\mathcal{F}} - U_{\mathcal{F}}^*| \le \int_{\mathcal{Y}} \sup_{f \in \mathcal{F}} |A(f,y) - A^*(f,y)| dy.$$
    Now, we bound the difference inside the supremum:
    $$|A(f, y) - A^*(f, y)| \le \int_{\mathcal{X}} |f(x)| |p(y|x) - p_*(y|x)| \mu(dx) + |p(y) - p_*(y)| \int_{\mathcal{X}} |f(x)| \mu(dx).$$
    We bound both terms uniformly using the growth profile $\omega_{\mathcal{F}}(x)$. Note that $\int_{\mathcal{X}} |f(x)|\mu(dx) \le \int_{\mathcal{X}} \omega_{\mathcal{F}}(x)\mu(dx) = \bar{\omega}$:
    $$|A(f, y) - A^*(f, y)| \le \int_{\mathcal{X}} \omega_{\mathcal{F}}(x) |p(y|x) - p_*(y|x)| \mu(dx) + \bar{\omega} |p(y) - p_*(y)|.$$
    Taking the supremum over $f$ simply yields this same upper bound. Integrating over $y$:
    $$|U_{\mathcal{F}} - U^*_{\mathcal{F}}| \le \int_{\mathcal{Y}} \int_{\mathcal{X}} \omega_{\mathcal{F}}(x) |p(y|x) - p_*(y|x)| \mu(dx) dy + \bar{\omega} \int_{\mathcal{Y}} |p(y) - p_*(y)| dy.$$
    Notice that the difference in evidences is bounded by the difference in likelihoods:
    $$\int_{\mathcal{Y}} |p(y) - p_*(y)| dy = \int_{\mathcal{Y}} \left| \int_{\mathcal{X}} (p(y|x) - p_*(y|x)) \mu(dx) \right| dy \le \int_{\mathcal{Y}} \int_{\mathcal{X}} |p(y|x) - p_*(y|x)| \mu(dx) dy.$$
    Substituting this into the previous inequality yields the desired bound:
    $$|U_{\mathcal{F}} - U^*_{\mathcal{F}}| \le \int_{\mathcal{Y}} \int_{\mathcal{X}} (\omega_{\mathcal{F}}(x) + \bar{\omega}) |p(y|x) - p_*(y|x)| \mu(dx) dy.$$
\end{proof}

\section{Proof of Prior Stability}
\label{sec:appendix_prior_proof}
\begin{proof}[Proof of Lemma \ref{lem:anchor}]
    Let $H(f) = \left| \int_{\mathcal{X}} f(x) p(dx) - \int_{\mathcal{X}} f(x) q(dx) \right|$. We wish to prove $\sup_{f \in \mathcal{F}} H(f) = \sup_{g \in \mathcal{F}_{\text{prior}}} H(g)$. Because $\mathcal{F}_{\text{prior}} \subset \mathcal{F}$, it trivially holds that $\sup_{f \in \mathcal{F}} H(f) \ge \sup_{g \in \mathcal{F}_{\text{prior}}} H(g)$. To prove the reverse inequality, consider any arbitrary function $f \in \mathcal{F}$. Let $c = \int_{\mathcal{X}} f(x) p(dx)$. By the shift-invariance of $\mathcal{F}$, the function $g(x) = f(x) - c$ belongs to $\mathcal{F}$. By construction, $\int_{\mathcal{X}} g(x) p(dx) = 0$, meaning that $g \in \mathcal{F}_{\text{prior}}$. We evaluate the posterior discrepancy score $H(g)$ for this shifted function:$$H(g) = \left| \int_{\mathcal{X}} (f(x) - c) p(dx) - \int_{\mathcal{X}} (f(x) - c) q(dx) \right|$$ 
    By the linearity of the integral, we can separate the constant $c$. Since $\mu^y$ and $\tilde{\mu}^y$ are both valid probability measures, they integrate to 1 (i.e., $\int d\mu^y = 1$ and $\int d\tilde{\mu}^y = 1$). The constants $c$ perfectly cancel. Hence, it holds that $$H(g) = \left| \int_{\mathcal{X}} f(x) p(dx) - \int_{\mathcal{X}} f(x) q^y(dx) \right| = H(f)$$
    Because every function $f \in \mathcal{F}$ has a corresponding shifted function $g \in \mathcal{F}_{\text{prior}}$ that yields the exact same absolute difference, the supremum over the full class cannot exceed the supremum over the restricted class. Thus, the supremums are strictly equal.
\end{proof}

\begin{proof}[Proof of Proposition \ref{prop:example}]
Let $\mathcal{X} = [0, R]$. We verify the pointwise existence of $C_E(y)$ and $C_L(y)$ for each metric, followed by their global $y$-integrability.

\textit{Part 1: Wasserstein-1 ($W_1$).} The dual class is $\mathcal{F}_W^{0} = \{f : f(0)=0, \|f'\|_\infty \le 1\}$.
The centered evidence function is $e_y(x) = \phi_\sigma(y-x) - \phi_\sigma(y-0)$. By definition, $e_y(0) = 0$. The minimal constant $C_E(y)$ such that $\frac{1}{C_E(y)} e_y \in \mathcal{F}_W^0$ is the Lipschitz constant of $e_y$. By the chain rule, $\frac{d}{dx} e_y(x) = -\phi_\sigma'(y-x)$, giving:
$$C_E(y) = \sup_{x \in [0,R]} \left| \frac{d}{dx} e_y(x) \right| = \sup_{x \in [0,R]} |\phi_\sigma'(y-x)|.$$
For the posterior constant $C_L(y)$, let $f \in \mathcal{F}_W^0$. We must bound the Lipschitz constant of $h(x) = f(x)\phi_\sigma(y-x)$. By the product rule:
$$|h'(x)| \le |f'(x)| \phi_\sigma(y-x) + |f(x)| |\phi_\sigma'(y-x)|.$$
Since $f \in \mathcal{F}_W^0$, $|f'(x)| \le 1$, we have $\|f\|_\infty \le R \|f'\|_\infty \le R$. Taking the supremum over $x \in [0,R]$ yields:
$$\|h'\|_\infty \le \sup_{x \in [0,R]} \phi_\sigma(y-x) + R \sup_{x \in [0,R]} |\phi_\sigma'(y-x)| = C_L(y).$$
Since both $C_E(y)$ and $C_L(y)$ are defined by the suprema of a Gaussian and its derivative over a compact interval, they exist pointwise for all $y$.

\textit{Part 2: Mat\'ern-MMD.} The dual class is defined strictly by the native norm: $\mathcal{F}_k = \{f \in \mathcal{H}_k : \|f\|_{\mathcal{H}_k} \le 1\}$.The uncentered evidence function is $e_y(x) = \phi_\sigma(y-x)$. Because the Gaussian is smooth, its restriction to $[0,R]$ belongs to $H^s([0,R])$ for any $s \ge 0$. To guarantee $\frac{1}{C_E(y)} e_y \in \mathcal{F}_k$, we require its $\mathcal{H}_k$ norm to be bounded by 1. Using the upper equivalence bound:$$\|e_y\|_{\mathcal{H}_k} \le c_2 \|e_y\|_{H^s([0,R])}.$$Thus, defining $C_E(y) = c_2 \|\phi_\sigma(y-\cdot)\|_{H^s([0,R])}$ satisfies the evidence condition. For the posterior constant $C_L(y)$, let $f \in \mathcal{F}_k$. By the lower equivalence bound, we know $\|f\|_{H^s} \le \frac{1}{c_1} \|f\|_{\mathcal{H}_k} \le \frac{1}{c_1}$. Because $s > 1/2$, $H^s([0,R])$ is a Banach algebra with multiplier constant $A_s > 0$. We bound the $\mathcal{H}_k$ norm of the product $f(\cdot)\phi_\sigma(y-\cdot)$ by chaining the upper equivalence, the Banach algebra property, and the lower equivalence:$$\|f \cdot e_y\|_{\mathcal{H}_k} \le c_2 \|f \cdot e_y\|_{H^s} \le c_2 A_s \|f\|_{H^s} \|e_y\|_{H^s} \le \frac{c_2 A_s}{c_1} \|e_y\|_{H^s}.$$We define $C_L(y) = \frac{c_2 A_s}{c_1} \|\phi_\sigma(y-\cdot)\|_{H^s([0,R])}$, which is strictly finite.

\textit{Part 3: Energy Distance.} The native space corresponding to the Euclidean energy distance in 1D, anchored at $0$, yields the dual class $\mathcal{F}_E^0 = \{f : f(0)=0, f \text{ absolutely continuous}, \|f'\|_{L^2} \le 1\}$. The centered evidence function is $e_y(x) = \phi_\sigma(y-x) - \phi_\sigma(y-0)$. We have 
$$C_E(y) = \left( \int_0^R \left( \frac{d}{dx} e_y(x) \right)^2 dx \right)^{1/2} = \left( \int_0^R (-\phi_\sigma'(y-x))^2 dx \right)^{1/2} = \|\phi_\sigma'(y-\cdot)\|_{L^2([0,R])}.$$
For $C_L(y)$, let $f \in \mathcal{F}_E^0$. The multiplier $h(x) = f(x)\phi_\sigma(y-x)$ must satisfy $h(0)=0$ and have a bounded $L^2$ derivative. By the product rule and Minkowski's inequality:$$\|h'\|_{L^2} = \|f' \phi_\sigma + f \phi_\sigma'\|_{L^2} \le \|f' \phi_\sigma\|_{L^2} + \|f \phi_\sigma'\|_{L^2}.$$We bound the terms separately. First, $\|f' \phi_\sigma\|_{L^2} \le \|f'\|_{L^2} \|\phi_\sigma(y-\cdot)\|_\infty \le 1 \cdot \sup_x \phi_\sigma(y-x)$. Second, because $f(0)=0$, Cauchy-Schwarz implies $|f(x)| = |\int_0^x f'(t)dt| \le \sqrt{x} \|f'\|_{L^2} \le \sqrt{R}$. Thus, $\|f \phi_\sigma'\|_{L^2} \le \sqrt{R} \|\phi_\sigma'(y-\cdot)\|_{L^2}$. Combining these gives:$$\|h'\|_{L^2} \le \sup_{x \in [0,R]} \phi_\sigma(y-x) + \sqrt{R} \|\phi_\sigma'(y-\cdot)\|_{L^2([0,R])} = C_L(y).$$

\textit{Part 4: Global Integrability in $y$.} For all three metrics, $C_E(y)$ and $C_L(y)$ are linear combinations of the supremum, $L^2$, or $H^s$ norms of $\phi_\sigma(y-\cdot)$ and its derivatives over $x \in [0,R]$.
As $|y| \to \infty$, the distance between $y$ and the compact set $[0,R]$ grows as $|y| - R$. Consequently, the magnitudes of $\phi_\sigma(y-x)$ and its spatial derivatives on $x \in [0,R]$ are uniformly bounded above by $K \exp\left( - \frac{(|y|-R)^2}{2\sigma^2} \right)$ for some polynomial $K(y)$.
Because the Gaussian tail decays faster than any polynomial, $C_E(y)$ and $C_L(y)$ decay exponentially as $|y| \to \infty$. Thus they are integrable.
\end{proof}

\begin{proof}[Proof of Theorem \ref{thm:prior}]
    By expanding the definitions of expected utility and applying the triangle inequality to the integrals over $\mathcal{Y}$, we can bound the difference by the supremum of the pointwise integrand errors:
    $$|U_{\mathcal{F}} - \tilde{U}_{\mathcal{F}}| \le \underbrace{\int_{\mathcal{Y}} p(y) \left| \gamma_{\mathcal{F}}(\mu, \mu^y) - \gamma_{\mathcal{F}}(\tilde{\mu}, \tilde{\mu}^y) \right| dy}_{\text{Term A}} + \underbrace{\int_{\mathcal{Y}} |p(y) - \tilde{p}(y)| \gamma_{\mathcal{F}}(\tilde{\mu}, \tilde{\mu}^y) dy}_{\text{Term B}}$$
    \textit{Bounding Term A:}
    By the reverse triangle inequality for metrics, $$|\gamma_{\mathcal{F}}(\mu, \mu^y) - \gamma_{\mathcal{F}}(\tilde{\mu}, \tilde{\mu}^y)| \le \gamma_{\mathcal{F}}(\mu, \tilde{\mu}) + \gamma_{\mathcal{F}}(\mu^y, \tilde{\mu}^y).$$ Multiplying by $p(y)$ and integrating:
    $$\text{Term A} \le \gamma_{\mathcal{F}}(\mu, \tilde{\mu}) + \int_{\mathcal{Y}} p(y) \gamma_{\mathcal{F}}(\mu^y, \tilde{\mu}^y) dy.$$
    Now we bound the expected difference between posteriors with IPM. Using the definition of the posterior and Lemma \ref{lem:anchor}:
    \begin{align*}
        p(y) \gamma_{\mathcal{F}}(\mu^y, \tilde{\mu}^y) &= \sup_{g \in \mathcal{F}_{\text{prior}}} \left| \int g(x) p(y|x) \mu(dx) - \frac{p(y)}{\tilde{p}(y)} \int g(x) p(y|x) \tilde{\mu}(dx) \right| \\
        & \le \sup_{f \in \mathcal{F}} \left| \int f(x) p(y|x) (\mu(dx) - \tilde{\mu}(dx)) \right|  \\
        & \quad + \sup_{g \in \mathcal{F}_{\text{prior}}} \left| \int g(x) p(y|x) \tilde{\mu}(dx) \right| \left| 1 - \frac{p(y)}{\tilde{p}(y)} \right|
    \end{align*}
    For the second part, note that $\frac{1}{\tilde{p}(y)} \int g(x)p(y|x)\tilde{\mu}(dx) = \int g d\tilde{\mu}^y$. The supremum of this within $\mathcal{F}_{\text{prior}}$ is exactly $D(y)$. For the first part, applying Assumption \ref{asmp:compatibility}, $f(\cdot)p(y|\cdot) = C_L(y) h(\cdot)$ for some $h \in \mathcal{F}$. Thus:$$\sup_{f \in \mathcal{F}} \left| \int f(x) p(y|x) (\mu(dx) - \tilde{\mu}(dx)) \right| \le C_L(y) \sup_{h \in \mathcal{F}} \left| \int h(x) (\mu(dx) - \tilde{\mu}(dx)) \right| = C_L(y) \gamma_{\mathcal{F}}(\mu, \tilde{\mu})$$
    Integrating these bounds for Term A over $y$:
    \begin{align*}\text{Term A} &\le \gamma_{\mathcal{F}}(\mu, \tilde{\mu}) + \int_{\mathcal{Y}} C_L(y) \gamma_{\mathcal{F}}(\mu, \tilde{\mu}) dy + \int_{\mathcal{Y}} D(y) |\tilde{p}(y) - p(y)|  dy \\
    & \le (1 + \int_{\mathcal{Y}} C_L(y) dy) \gamma_{\mathcal{F}}(\mu, \tilde{\mu}) + \int_{\mathcal{Y}} D(y) |p(y) - \tilde{p}(y)| dy
    \end{align*}
    \textit{Bounding Term B:} According to Lemma \ref{lem:anchor}, we have:
    $$\text{Term B} \le \int_{\mathcal{Y}} D(y)|p(y) - \tilde{p}(y)| dy$$
    Adding Term A and Term B gives
    $$ |U_{\mathcal{F}} - \tilde{U}_{\mathcal{F}}| \le (1 + \bar{C}_L) \gamma_{\mathcal{F}}(\mu, \tilde{\mu}) + 2 \int_{\mathcal{Y}} D(y)|p(y) - \tilde{p}(y)| dy $$
    Again, we use Assumption \ref{asmp:compatibility}
    \begin{align*}
    |p(y) - \tilde{p}(y)| &= \left| \int p(y|x) \mu(dx) - \int p(y|x) \tilde{\mu}(dx) \right| \\
    &=\left| \int e_y(x) \mu(dx) - \int e_y(x) \tilde{\mu}(dx) \right| \\
    &\le C_E(y) \sup_{h \in \mathcal{F}} \left| \int h(x) \mu(dx) - \int h(x) \tilde{\mu}(dx) \right| \\
    &= C_E(y) \gamma_{\mathcal{F}}(\mu, \tilde{\mu}).
    \end{align*}
    Now, integrate this over all observations $y$:
    $$\int_{\mathcal{Y}} D(y)|p(y) - \tilde{p}(y)| dy \le \left( \int_{\mathcal{Y}} C_E(y)D(y) dy \right) \gamma_{\mathcal{F}}(\mu, \tilde{\mu}).$$
    This gives the final bound. 
\end{proof}

\section{Rigorous Discussion for Section \ref{sec:understanding}}
\label{sec:math_understanding}
\paragraph{Likelihood Stability} Consider the observation is generated by a forward model with Gaussian noise: $y = G(x) + \eta$, where $\eta \sim \mathcal{N}(0, \Sigma)$. The true data misfit potential is:$$\Phi(x; y) = \frac{1}{2} \big( y - G(x) \big)^\top \Sigma^{-1} \big( y - G(x) \big)$$The true likelihood is $p(y|x) = \frac{1}{Z}\exp(-\Phi(x;y))$. Let the perturbed model be $G_*(x) = G(x) + \delta(x)$. Assuming the modeling error scales with the model magnitude, the perturbation is bounded as $\|\delta(x)\| \le \epsilon(1 + \|G(x)\|)$. The difference in potentials is bounded by:
$$
\begin{aligned}
    |\Delta \Phi| = & \left| \Phi_*(x; y) - \Phi(x; y) \right| \\
    = & \left| - \big( y - G(x) \big)^\top \Sigma^{-1} \delta(x) + \frac{1}{2} \delta(x)^\top \Sigma^{-1} \delta(x) \right| \\
    \le &  \epsilon (1 + \|G(x)\|) \|\Sigma^{-1}\| \|y - G(x)\| + \mathcal{O}(\epsilon^2).
\end{aligned}
$$
By the Mean Value Theorem, for a sufficiently small model perturbation $\epsilon \to 0$, the second-order term vanishes, and we obtain the first-order bound on the likelihoods:$$\left| p_*(y|x) - p(y|x) \right| \le p(y|x) \cdot \exp(|\Delta \Phi|) \cdot |\Delta \Phi| \lesssim p(y|x) \cdot \Big[ \epsilon (1 + \|G(x)\|) \|\Sigma^{-1}\| \|y - G(x)\| \Big].$$

Substituting this into the general expected utility bound yields:

$$|U_{\mathcal{F}} - U_{\mathcal{F}}^*| \lesssim \epsilon \|\Sigma^{-1}\| \int_{\mathcal{X}} \omega_{\mathcal{F}}(x) (1 + \|G(x)\|) \left( \int_{\mathcal{Y}} \|y - G(x)\| p(y|x) dy \right) \mu(dx).$$Notice that the inner integral over $\mathcal{Y}$ is simply the expected magnitude of the Gaussian noise $\mathbb{E}[\|\eta\|]$, which evaluates to a constant $K$. The bound simplifies to:$$|U_{\mathcal{F}} - U_{\mathcal{F}}^*| \lesssim \epsilon K \|\Sigma^{-1}\| \int_{\mathcal{X}} \omega_{\mathcal{F}}(x) (1 + \|G(x)\|) \mu(dx).$$

If we assume $G(x)$ is globally Lipschitz ($L_1$), then $\|G(x)\| \le L_1\|x\| + \|G(0)\|$. The polynomial order of the integrand dictates the prior moment requirements:
\begin{itemize}
    \item MMD ($\omega_{\mathcal{F}}(x) = \mathcal{O}(1)$): Integrand is $\mathcal{O}(\|x\|) \implies$ A finite 1st moment is sufficient.
    \item Energy Distance ($\omega_{\mathcal{F}}(x) = \mathcal{O}(\|x\|^{1/2})$): Integrand is $\mathcal{O}(\|x\|^{1.5}) \implies$ A finite 1.5-th moment is sufficient.
    \item Wasserstein-1 ($\omega_{\mathcal{F}}(x) = \mathcal{O}(\|x\|)$): Integrand is $\mathcal{O}(\|x\|^2) \implies$ A finite 2nd moment is sufficient.
\end{itemize}

\paragraph{Prior Stability} The posterior discrepancy is defined by the dual representation:
$$D(y) = \gamma_{\mathcal{F}}(\mu^y, \tilde{\mu}^y) = \sup_{f \in \mathcal{F}} \left| \int_{\mathcal{X}} f(x) \mu^y(dx) - \int_{\mathcal{X}} f(x) \tilde{\mu}^y(dx) \right|.$$

For bounded metrics (MMD), $\sup_{f \in \mathcal{F}} |f(x)| \le M$. The difference is trivially bounded by $2M$, eliminating the need to analyze spatial gradients or invoke transport bounds.

For unbounded metrics ($W_1$, Energy), note that the posterior is $$\mu^y(dx) = \frac{1}{Z} \exp(-\Phi(x;y)) \mu(dx)$$
where $Z$ is the marginal likelihood. Then 
$$
    \begin{aligned}
        & \int_{\mathcal{X}} f(x) \mu^y(dx) - \int_{\mathcal{X}} f(x) \tilde{\mu}^y(dx) \\ 
        = & \frac{1}{Z} \int_{\mathcal{X}} f(x) \exp(-\Phi(x;y)) \mu(dx) - \frac{1}{\tilde{Z}} \int_{\mathcal{X}} f(x) \exp(-\Phi(x;y)) \tilde{\mu}(dx)
    \end{aligned}
$$
For convenience, let $I$ and $\tilde{I}$ represent those unnormalized integrals:
$$\int_{\mathcal{X}} f(x) \mu^y(dx) - \int_{\mathcal{X}} f(x) \tilde{\mu}^y(dx)= \frac{I}{Z} - \frac{\tilde{I}}{\tilde{Z}}= \frac{1}{Z} \big( I - \tilde{I} \big) + \frac{\tilde{I}}{Z \tilde{Z}} \big( \tilde{Z} - Z \big)$$

We denote the first term $(I - \tilde{I})$ as $\Delta I(y)$. The second term $(\tilde{Z} - Z)$ is the difference in unnormalized integrals where the test function is a constant $f(x) = 1$ (Note that the multiplier $|\tilde{I}|$ in the second term is trivially bounded by the prior's first moment $\int \omega_{\mathcal{F}}(x) \tilde{\mu}(dx)$, which is guaranteed to be finite under the higher-order moment assumptions required to bound $\Delta I(f)$). Then the core mathematical challenge lies in bounding the difference in the unnormalized integrals:$$\Delta I(y) = \left| \int_{\mathcal{X}} f(x) \exp(-\Phi(x;y)) \mu(dx) - \int_{\mathcal{X}} f(x) \exp(-\Phi(x;y)) \tilde{\mu}(dx) \right|.$$Let $h_y(x) = f(x) \exp(-\Phi(x;y))$ denote the effective integrand. We can rewrite the difference using any joint coupling $p(x, x_*)$ whose marginals are $\mu$ and $\tilde{\mu}$:
$$\Delta I(y) = \left| \int_{\mathcal{X} \times \mathcal{X}} \big( h_y(x) - h_y(x_*) \big) p(dx, dx_*) \right|.$$
By the Mean Value Theorem, for every pair $(x, x_*)$, there exists a point $z$ on the line segment between them such that:$$|h_y(x) - h_y(x_*)| \le \|\nabla h_y(z)\| \|x - x_*\|.$$
This yields the transport bound:$$\Delta I(y) \le \int_{\mathcal{X} \times \mathcal{X}} \|\nabla h_y(z)\| \|x - x_*\| p(dx, dx_*).$$
Applying the product rule, its gradient is:
$$\nabla h_y(x) = \nabla f(x) \exp(-\Phi(x;y)) - f(x) \exp(-\Phi(x;y)) \nabla_x \Phi(x;y)$$
Since $\exp(-\Phi) \le 1$, the norm is strictly bounded by:
$$\|\nabla h_y(x)\| \le \|\nabla f(x)\| + |f(x)| \|\nabla_x \Phi(x;y)\| \le \|\nabla f(x)\| + \omega_{\mathcal{F}}(x) \|\nabla_x \Phi(x;y)\|.$$
For a Gaussian observation model $y = G(x) + \eta$ with noise covariance $\Sigma$, the gradient of the data misfit potential is $\nabla_x \Phi(x;y) = -J_G(x)^\top \Sigma^{-1} \big(y - G(x)\big)$. If we apply the Lipschitz assumption for the forward model ($L_1$), we have $\|J_G(x)\| \le L_1$ and $\|G(x)\| \le L_1\|x\| + \|G(0)\|$. Substituting these into the gradient bound yields:
$$
\begin{aligned}
    \|\nabla_x \Phi(x;y)\| &\le L_1 \|\Sigma^{-1}\| \big( \|y\| + \|G(x)\| \big) \\
    &\le L_1 \|\Sigma^{-1}\| \big( L_1\|x\| + \|y\| + \|G(0)\| \big) \\
    &= \mathcal{O}(L_1^2 \|x\|) + C_y.
\end{aligned}
$$

Because $\|\nabla f(x)\|$ is locally integrable or strictly bounded depending on the metric, the tail behavior as $\|x\| \to \infty$ is strictly dominated by the product $\omega_{\mathcal{F}}(x) \|\nabla_x \Phi(x;y)\|$:

\begin{itemize}
    \item \textbf{Energy Distance:} With the energy distance, the test function envelope grows sub-linearly, giving us $\omega_E(x) = \mathcal{O}(\|x\|^{1/2})$. Because of this, the gradient scales as:
    $$\|\nabla h_y(x)\| \approx \mathcal{O}(\|x\|^{1/2}) \times \mathcal{O}(L_1^2 \|x\|) = \mathcal{O}(L_1^2 \|x\|^{1.5}).$$
    When we plug this into the transport bound $\int \|\nabla h_y(z)\| \|x - x_*\| p(dx, dx_*)$, the resulting integrand scales as $\mathcal{O}(\|x\|^{2.5})$. In practice, this means the prior only needs a finite 2.5-th moment to keep the discrepancy finite, which gives us the flexibility to use heavy-tailed priors with polynomial decay.
    
    \item \textbf{Wasserstein-1:} For Wasserstein-1, the test function envelope grows linearly, meaning $\omega_{W_1}(x) = \mathcal{O}(\|x\|)$. This pushes the gradient scaling to a quadratic rate:
    $$\|\nabla h_y(x)\| \approx \mathcal{O}(\|x\|) \times \mathcal{O}(L_1^2 \|x\|) = \mathcal{O}(L_1^2 \|x\|^2).$$
    As a result, the transport integrand scales as $\mathcal{O}(\|x\|^3)$. However, the real constraint comes from the overarching expected utility bound, which requires us to integrate $D(y)$ over all possible observations $y \in \mathcal{Y}$. Under extreme observation shifts, the polynomial warping inside $D(y)$ scales with $\exp(\mathcal{O}(L_1^2 \|x\|^2))$. To keep this bound firmly under control, we are forced to use a prior with exponential tail decay. This naturally leads to the standard assumption of a strictly sub-Gaussian prior $\mu(x) \propto \exp(-L_2\|x\|^2)$, where the geometric condition $L_1^2 < C L_2$ ensures the prior decays fast enough to overpower the forward model's warping.
\end{itemize}

\vskip 0.2in
\bibliography{sample}

@article{foster2019variational,
  title={Variational {Bayesian} optimal experimental design},
  author={Foster, Adam and Jankowiak, Martin and Bingham, Elias and Horsfall, Paul and Teh, Yee Whye and Rainforth, Thomas and Goodman, Noah},
  journal={Advances in Neural Information Processing Systems},
  volume={32},
  year={2019}
}

@inproceedings{rainforth2018nesting,
  title={On nesting {Monte} {Carlo} estimators},
  author={Rainforth, Tom and Cornish, Rob and Yang, Hongseok and Warrington, Andrew and Wood, Frank},
  booktitle={International Conference on Machine Learning},
  pages={4267--4276},
  year={2018},
  organization={PMLR}
}

@article{barber2004algorithm,
  title={The {IM} algorithm: {A} variational approach to information maximization},
  author={Barber, David and Agakov, Felix},
  journal={Advances in Neural Information Processing Systems},
  volume={16},
  number={320},
  pages={201},
  year={2004}
}

@article{helin2025bayesian,
  title={Bayesian optimal experimental design with {Wasserstein} information criteria},
  author={Helin, Tapio and Marzouk, Youssef and Rojo-Garcia, Jose Rodrigo},
  journal={arXiv preprint arXiv:2504.10092},
  year={2025}
}

@inproceedings{makkuva2020optimal,
  title={Optimal transport mapping via input convex neural networks},
  author={Makkuva, Ashok and Taghvaei, Amirhossein and Oh, Sewoong and Lee, Jason},
  booktitle={International Conference on Machine Learning},
  pages={6672--6681},
  year={2020},
  organization={PMLR}
}

@article{lindley1956measure,
  title={On a measure of the information provided by an experiment},
  author={Lindley, Dennis V},
  journal={The Annals of Mathematical Statistics},
  volume={27},
  number={4},
  pages={986--1005},
  year={1956},
  publisher={Institute of Mathematical Statistics}
}

@article{chaloner1995bayesian,
  title={Bayesian experimental design: A review},
  author={Chaloner, Kathryn and Verdinelli, Isabella},
  journal={Statistical Science},
  pages={273--304},
  year={1995},
  publisher={JSTOR}
}

@article{shannon1948mathematical,
  title={A mathematical theory of communication},
  author={Shannon, Claude Elwood},
  journal={The Bell System Technical Journal},
  volume={27},
  number={3},
  pages={379--423},
  year={1948},
  publisher={Nokia Bell Labs}
}

@article{ryan2016review,
  title={A review of modern computational algorithms for {Bayesian} optimal design},
  author={Ryan, Elizabeth G and Drovandi, Christopher C and McGree, James M and Pettitt, Anthony N},
  journal={International Statistical Review},
  volume={84},
  number={1},
  pages={128--154},
  year={2016},
  publisher={Wiley Online Library}
}

@article{sebastiani2000maximum,
  title={Maximum entropy sampling and optimal {Bayesian} experimental design},
  author={Sebastiani, Paola and Wynn, Henry P},
  journal={Journal of the Royal Statistical Society: Series B (Statistical Methodology)},
  volume={62},
  number={1},
  pages={145--157},
  year={2000},
  publisher={Wiley Online Library}
}

@article{bernardo1979expected,
  title={Expected information as expected utility},
  author={Bernardo, Jos{\'e} M},
  journal={The Annals of Statistics},
  pages={686--690},
  year={1979},
  publisher={JSTOR}
}

@article{walker2016bayesian,
  title={Bayesian information in an experiment and the {Fisher} information distance},
  author={Walker, Stephen G},
  journal={Statistics \& Probability Letters},
  volume={112},
  pages={5--9},
  year={2016},
  publisher={Elsevier}
}

@article{overstall2022properties,
  title={Properties of {Fisher} information gain for {Bayesian} design of experiments},
  author={Overstall, Antony M},
  journal={Journal of Statistical Planning and Inference},
  volume={218},
  pages={138--146},
  year={2022},
  publisher={Elsevier}
}

@article{prangle2023bayesian,
  title={Bayesian experimental design without posterior calculations: {An} adversarial approach},
  author={Prangle, Dennis and Harbisher, Sophie and Gillespie, Colin S},
  journal={Bayesian Analysis},
  volume={18},
  number={1},
  pages={133--163},
  year={2023},
  publisher={International Society for Bayesian Analysis}
}

@article{long2013fast,
  title={Fast estimation of expected information gains for {Bayesian} experimental designs based on {Laplace} approximations},
  author={Long, Quan and Scavino, Marco and Tempone, Ra{\'u}l and Wang, Suojin},
  journal={Computer Methods in Applied Mechanics and Engineering},
  volume={259},
  pages={24--39},
  year={2013},
  publisher={Elsevier}
}

@article{beck2018fast,
  title={Fast {Bayesian} experimental design: {Laplace}-based importance sampling for the expected information gain},
  author={Beck, Joakim and Dia, Ben Mansour and Espath, Luis FR and Long, Quan and Tempone, Raul},
  journal={Computer Methods in Applied Mechanics and Engineering},
  volume={334},
  pages={523--553},
  year={2018},
  publisher={Elsevier}
}

@article{myung2013tutorial,
  title={A tutorial on adaptive design optimization},
  author={Myung, Jay I and Cavagnaro, Daniel R and Pitt, Mark A},
  journal={Journal of Mathematical Psychology},
  volume={57},
  number={3-4},
  pages={53--67},
  year={2013},
  publisher={Elsevier}
}

@inproceedings{foster2021deep,
  title={Deep adaptive design: {Amortizing} sequential bayesian experimental design},
  author={Foster, Adam and Ivanova, Desi R and Malik, Ilyas and Rainforth, Tom},
  booktitle={International Conference on Machine Learning},
  pages={3384--3395},
  year={2021},
  organization={PMLR}
}

@article{huan2013simulation,
  title={Simulation-based optimal {Bayesian} experimental design for nonlinear systems},
  author={Huan, Xun and Marzouk, Youssef M},
  journal={Journal of Computational Physics},
  volume={232},
  number={1},
  pages={288--317},
  year={2013},
  publisher={Elsevier}
}

@inproceedings{kleinegesse2019efficient,
  title={Efficient Bayesian experimental design for implicit models},
  author={Kleinegesse, Steven and Gutmann, Michael U},
  booktitle={The 22nd International Conference on Artificial Intelligence and Statistics},
  pages={476--485},
  year={2019},
  organization={PMLR}
}

@article{shewry1987maximum,
  title={Maximum entropy sampling},
  author={Shewry, Michael C and Wynn, Henry P},
  journal={Journal of Applied Statistics},
  volume={14},
  number={2},
  pages={165--170},
  year={1987},
  publisher={Taylor \& Francis}
}

@article{rainforth2024modern,
  title={Modern {Bayesian} experimental design},
  author={Rainforth, Tom and Foster, Adam and Ivanova, Desi R and Bickford Smith, Freddie},
  journal={Statistical Science},
  volume={39},
  number={1},
  pages={100--114},
  year={2024},
  publisher={Institute of Mathematical Statistics}
}

@article{huan2024optimal,
  title={Optimal experimental design: Formulations and computations},
  author={Huan, Xun and Jagalur, Jayanth and Marzouk, Youssef},
  journal={Acta Numerica},
  volume={33},
  pages={715--840},
  year={2024},
  publisher={Cambridge University Press}
}

@inproceedings{arjovsky2017wasserstein,
  title={Wasserstein generative adversarial networks},
  author={Arjovsky, Martin and Chintala, Soumith and Bottou, L{\'e}on},
  booktitle={International Conference on Machine Learning},
  pages={214--223},
  year={2017},
  organization={PMLR}
}

@article{kerrigan2025geometric,
  title={A Geometric Approach to Optimal Experimental Design},
  author={Kerrigan, Gavin and Naesseth, Christian A and Rainforth, Tom},
  journal={arXiv preprint arXiv:2510.14848},
  year={2025}
}

@book{peyre2019computational,
  title={Computational optimal transport: With applications to data science},
  author={Peyr{\'e}, Gabriel and Cuturi, Marco},
  year={2019},
  publisher={Now Foundations and Trends}
}

@article{wu2025pins,
  title={{PINS}: {Proximal} Iterations with Sparse {Newton} and {Sinkhorn} for Optimal Transport},
  author={Wu, Di and Liang, Ling and Yang, Haizhao},
  journal={arXiv preprint arXiv:2502.03749},
  year={2025}
}

@book{villani2009optimal,
  title={Optimal transport: {Old} and new},
  author={Villani, C{\'e}dric and others},
  volume={338},
  year={2009},
  publisher={Springer}
}

@article{muller1997integral,
  title={Integral probability metrics and their generating classes of functions},
  author={M{\"u}ller, Alfred},
  journal={Advances in Applied Probability},
  volume={29},
  number={2},
  pages={429--443},
  year={1997},
  publisher={Cambridge University Press}
}

@inproceedings{amos2017input,
  title={Input convex neural networks},
  author={Amos, Brandon and Xu, Lei and Kolter, J Zico},
  booktitle={International Conference on Machine Learning},
  pages={146--155},
  year={2017},
  organization={PMLR}
}

@article{korotin2019wasserstein,
  title={Wasserstein-2 generative networks},
  author={Korotin, Alexander and Egiazarian, Vage and Asadulaev, Arip and Safin, Alexander and Burnaev, Evgeny},
  journal={arXiv preprint arXiv:1909.13082},
  year={2019}
}

@article{lookman2019active,
  title={Active learning in materials science with emphasis on adaptive sampling using uncertainties for targeted design},
  author={Lookman, Turab and Balachandran, Prasanna V and Xue, Dezhen and Yuan, Ruihao},
  journal={npj Computational Materials},
  volume={5},
  number={1},
  pages={21},
  year={2019},
  publisher={Nature Publishing Group UK London}
}

@article{sacks1989design,
  title={Design and analysis of computer experiments},
  author={Sacks, Jerome and Welch, William J and Mitchell, Toby J and Wynn, Henry P},
  journal={Statistical Science},
  volume={4},
  number={4},
  pages={409--423},
  year={1989},
  publisher={Institute of Mathematical Statistics}
}

@article{hellmuth2025data,
  title={Data selection: at the interface of {PDE}-based inverse problem and randomized linear algebra},
  author={Hellmuth, Kathrin and Jin, Ruhui and Li, Qin and Wright, Stephen J},
  journal={arXiv preprint arXiv:2510.01567},
  year={2025}
}

@article{jin2024continuous,
  title={Continuous nonlinear adaptive experimental design with gradient flow},
  author={Jin, Ruhui and Li, Qin and Mussmann, Stephen O and Wright, Stephen J},
  journal={arXiv preprint arXiv:2411.14332},
  year={2024}
}

@article{chen2026data,
  title={Data Completion for Electrical Impedance Tomography by Conditional Diffusion Models},
  author={Chen, Ke and Yang, Haizhao and Yi, Chugang},
  journal={arXiv preprint arXiv:2602.07813},
  year={2026}
}

@article{haber2008numerical,
  title={Numerical methods for experimental design of large-scale linear ill-posed inverse problems},
  author={Haber, Eldad and Horesh, Lior and Tenorio, Luis},
  journal={Inverse Problems},
  volume={24},
  number={5},
  pages={055012},
  year={2008}
}

@article{alexanderian2016fast,
  title={A fast and scalable method for {A}-optimal design of experiments for infinite-dimensional {Bayesian} nonlinear inverse problems},
  author={Alexanderian, Alen and Petra, Noemi and Stadler, Georg and Ghattas, Omar},
  journal={SIAM Journal on Scientific Computing},
  volume={38},
  number={1},
  pages={A243--A272},
  year={2016},
  publisher={SIAM}
}

@article{ruthotto2018optimal,
  title={Optimal experimental design for inverse problems with state constraints},
  author={Ruthotto, Lars and Chung, Julianne and Chung, Matthias},
  journal={SIAM Journal on Scientific Computing},
  volume={40},
  number={4},
  pages={B1080--B1100},
  year={2018},
  publisher={SIAM}
}

@article{alexanderian2021optimal,
  title={Optimal experimental design for infinite-dimensional {Bayesian} inverse problems governed by {PDEs}: {A} review},
  author={Alexanderian, Alen},
  journal={Inverse Problems},
  volume={37},
  number={4},
  pages={043001},
  year={2021},
  publisher={IOP Publishing}
}

@article{helin2022edge,
  title={Edge-promoting adaptive {Bayesian} experimental design for {X}-ray imaging},
  author={Helin, Tapio and Hyvonen, Nuutti and Puska, Juha-Pekka},
  journal={SIAM Journal on Scientific Computing},
  volume={44},
  number={3},
  pages={B506--B530},
  year={2022},
  publisher={SIAM}
}

@article{steinberg1984experimental,
  title={Experimental design: {Review} and comment},
  author={Steinberg, David M and Hunter, William G},
  journal={Technometrics},
  volume={26},
  number={2},
  pages={71--97},
  year={1984},
  publisher={Taylor \& Francis}
}

@book{dudley2018real,
  title={Real analysis and probability},
  author={Dudley, Richard M},
  year={2018},
  publisher={Chapman and Hall/CRC}
}

@book{berlinet2011reproducing,
  title={Reproducing kernel Hilbert spaces in probability and statistics},
  author={Berlinet, Alain and Thomas-Agnan, Christine},
  year={2011},
  publisher={Springer Science \& Business Media}
}

@article{gretton2012kernel,
  title={A kernel two-sample test},
  author={Gretton, Arthur and Borgwardt, Karsten M and Rasch, Malte J and Sch{\"o}lkopf, Bernhard and Smola, Alexander},
  journal={The Journal of Machine Learning Research},
  volume={13},
  number={1},
  pages={723--773},
  year={2012},
  publisher={JMLR. org}
}

@article{szekely2004testing,
  title={Testing for equal distributions in high dimension},
  author={Sz{\'e}kely, G{\'a}bor J and Rizzo, Maria L and others},
  journal={InterStat},
  volume={5},
  number={16.10},
  pages={1249--1272},
  year={2004}
}

@article{szekely2005new,
  title={A new test for multivariate normality},
  author={Sz{\'e}kely, G{\'a}bor J and Rizzo, Maria L},
  journal={Journal of Multivariate Analysis},
  volume={93},
  number={1},
  pages={58--80},
  year={2005},
  publisher={Elsevier}
}

@article{gretton2009fast,
  title={A fast, consistent kernel two-sample test},
  author={Gretton, Arthur and Fukumizu, Kenji and Harchaoui, Zaid and Sriperumbudur, Bharath K},
  journal={Advances in Neural Information Processing Systems},
  volume={22},
  year={2009}
}

@inproceedings{huadaptive,
  title={Adaptive Defense against Harmful Fine-Tuning for Large Language Models via {Bayesian} Data Scheduler},
  author={Hu, Zixuan and Shen, Li and Wang, Zhenyi and Wei, Yongxian and Tao, Dacheng},
  booktitle={The Thirty-ninth Annual Conference on Neural Information Processing Systems},
  year={2025}
}

@article{chang2025provable,
  title={Provable diffusion posterior sampling for {Bayesian} inversion},
  author={Chang, Jinyuan and Duan, Chenguang and Jiao, Yuling and Li, Ruoxuan and Yang, Jerry Zhijian and Yuan, Cheng},
  journal={arXiv preprint arXiv:2512.08022},
  year={2025}
}

@article{cang2022supervised,
  title={Supervised optimal transport},
  author={Cang, Zixuan and Nie, Qing and Zhao, Yanxiang},
  journal={SIAM Journal on Applied Mathematics},
  volume={82},
  number={5},
  pages={1851--1877},
  year={2022},
  publisher={SIAM}
}

@article{dunbar2022ensemble,
  title={Ensemble-based experimental design for targeting data acquisition to inform climate models},
  author={Dunbar, Oliver RA and Howland, Michael F and Schneider, Tapio and Stuart, Andrew M},
  journal={Journal of Advances in Modeling Earth Systems},
  volume={14},
  number={9},
  pages={e2022MS002997},
  year={2022},
  publisher={Wiley Online Library}
}

@inproceedings{bhatt2024experimental,
  title={An experimental design framework for label-efficient supervised finetuning of large language models},
  author={Bhatt, Gantavya and Chen, Yifang and Das, Arnav and Zhang, Jifan and Truong, Sang and Mussmann, Stephen and Zhu, Yinglun and Bilmes, Jeff and Du, Simon and Jamieson, Kevin and others},
  booktitle={Findings of the Association for Computational Linguistics: ACL 2024},
  pages={6549--6560},
  year={2024}
}

@article{fiez2019sequential,
  title={Sequential experimental design for transductive linear bandits},
  author={Fiez, Tanner and Jain, Lalit and Jamieson, Kevin G and Ratliff, Lillian},
  journal={Advances in Neural Information Processing Systems},
  volume={32},
  year={2019}
}

@inproceedings{camilleri2021high,
  title={High-dimensional experimental design and kernel bandits},
  author={Camilleri, Romain and Jamieson, Kevin and Katz-Samuels, Julian},
  booktitle={International Conference on Machine Learning},
  pages={1227--1237},
  year={2021},
  organization={PMLR}
}

@article{wagenmaker2022instance,
  title={Instance-dependent near-optimal policy identification in linear {MDPs} via online experiment design},
  author={Wagenmaker, Andrew and Jamieson, Kevin G},
  journal={Advances in Neural Information Processing Systems},
  volume={35},
  pages={5968--5981},
  year={2022}
}

@article{chen2015selection,
  title={Selection and estimation for mixed graphical models},
  author={Chen, Shizhe and Witten, Daniela M and Shojaie, Ali},
  journal={Biometrika},
  volume={102},
  number={1},
  pages={47--64},
  year={2015},
  publisher={Oxford University Press}
}

@article{dror2008sequential,
  title={Sequential experimental designs for generalized linear models},
  author={Dror, Hovav A and Steinberg, David M},
  journal={Journal of the American Statistical Association},
  volume={103},
  number={481},
  pages={288--298},
  year={2008},
  publisher={Taylor \& Francis}
}

@article{mackay1992information,
  title={Information-based objective functions for active data selection},
  author={MacKay, David JC},
  journal={Neural computation},
  volume={4},
  number={4},
  pages={590--604},
  year={1992},
  publisher={MIT Press One Rogers Street, Cambridge, MA 02142-1209, USA journals-info~…}
}

@article{ho2020denoising,
  title={Denoising diffusion probabilistic models},
  author={Ho, Jonathan and Jain, Ajay and Abbeel, Pieter},
  journal={Advances in Neural Information Processing Systems},
  volume={33},
  pages={6840--6851},
  year={2020}
}

@article{lipman2022flow,
  title={Flow matching for generative modeling},
  author={Lipman, Yaron and Chen, Ricky TQ and Ben-Hamu, Heli and Nickel, Maximilian and Le, Matt},
  journal={arXiv preprint arXiv:2210.02747},
  year={2022}
}

@article{he2008active,
  title={Active learning of causal networks with intervention experiments and optimal designs},
  author={He, Yang-Bo and Geng, Zhi},
  journal={Journal of Machine Learning Research},
  volume={9},
  number={11},
  year={2008}
}

@article{zhu2025balancing,
  title={Balancing Interference and Correlation in Spatial Experimental Designs: A Causal Graph Cut Approach},
  author={Zhu, Jin and Li, Jingyi and Zhou, Hongyi and Lin, Yinan and Lin, Zhenhua and Shi, Chengchun},
  journal={arXiv preprint arXiv:2505.20130},
  year={2025}
}

@article{liu2023review,
  title={A review of recent advances in empirical likelihood},
  author={Liu, Pangpang and Zhao, Yichuan},
  journal={Wiley Interdisciplinary Reviews: Computational Statistics},
  volume={15},
  number={3},
  pages={e1599},
  year={2023},
  publisher={Wiley Online Library}
}

@article{sun2024arma,
  title={Arma-design: Optimal treatment allocation strategies for a/b testing in partially observable time series experiments},
  author={Sun, Ke and Kong, Linglong and Zhu, Hongtu and Shi, Chengchun},
  journal={arXiv preprint arXiv:2408.05342},
  year={2024}
}

@article{song2026wasserstein,
  title={Wasserstein generative regression},
  author={Song, Shanshan and Wang, Tong and Shen, Guohao and Lin, Yuanyuan and Huang, Jian},
  journal={Journal of the Royal Statistical Society Series B: Statistical Methodology},
  volume={88},
  number={1},
  pages={330--351},
  year={2026},
  publisher={Oxford University Press UK}
}

@article{liu2021wasserstein,
  title={Wasserstein generative learning of conditional distribution},
  author={Liu, Shiao and Zhou, Xingyu and Jiao, Yuling and Huang, Jian},
  journal={arXiv preprint arXiv:2112.10039},
  year={2021}
}

@article{liang2024pnod,
  title={{PNOD}: An Efficient Projected {Newton} Framework for Exact Optimal Experimental Designs},
  author={Liang, Ling and Yang, Haizhao},
  journal={https://arxiv.org/abs/2409.18392},
  year={2024}
}

@inproceedings{smola2007hilbert,
  title={A {Hilbert} space embedding for distributions},
  author={Smola, Alex and Gretton, Arthur and Song, Le and Sch{\"o}lkopf, Bernhard},
  booktitle={International Conference on Algorithmic Learning Theory},
  pages={13--31},
  year={2007},
  organization={Springer}
}

@article{gao2026observation,
  title={Observationally Informed Adaptive Causal Experimental Design},
  author={Gao, Erdun and Zhang, Liang and Fawkes, Jake and Zuo, Aoqi and Liu, Wenqin and Li, Haoxuan and Gong, Mingming and Sejdinovic, Dino},
  journal={arXiv preprint arXiv:2603.03785},
  year={2026}
}

@article{sejdinovic2013equivalence,
  title={Equivalence of distance-based and {RKHS-based} statistics in hypothesis testing},
  author={Sejdinovic, Dino and Sriperumbudur, Bharath and Gretton, Arthur and Fukumizu, Kenji},
  journal={The Annals of Statistics},
  pages={2263--2291},
  year={2013},
  publisher={JSTOR}
}

@article{hanin1992kantorovich,
  title={Kantorovich-{Rubinstein} norm and its application in the theory of {Lipschitz} spaces},
  author={Hanin, Leonid G},
  journal={Proceedings of the American Mathematical Society},
  volume={115},
  number={2},
  pages={345--352},
  year={1992}
}

@article{tolstikhin2017minimax,
  title={Minimax estimation of kernel mean embeddings},
  author={Tolstikhin, Ilya and Sriperumbudur, Bharath K and Muandet, Krikamol},
  journal={Journal of Machine Learning Research},
  volume={18},
  number={86},
  pages={1--47},
  year={2017}
}

@article{di2012hitchhiker,
  title={Hitchhiker's guide to the fractional {Sobolev} spaces},
  author={Di Nezza, Eleonora and Palatucci, Giampiero and Valdinoci, Enrico},
  journal={Bulletin Des Sciences Math{\'e}matiques},
  volume={136},
  number={5},
  pages={521--573},
  year={2012},
  publisher={Elsevier}
}

@article{strichartz1967multipliers,
  title={Multipliers on fractional {Sobolev} spaces},
  author={Strichartz, Robert S},
  journal={Journal of Mathematics and Mechanics},
  volume={16},
  number={9},
  pages={1031--1060},
  year={1967},
  publisher={JSTOR}
}

@article{grafakos2014kato,
  title={The kato-ponce inequality},
  author={Grafakos, Loukas and Oh, Seungly},
  journal={Communications in Partial Differential Equations},
  volume={39},
  number={6},
  pages={1128--1157},
  year={2014},
  publisher={Taylor \& Francis}
}

@article{brasco2021characterisation,
  title={Characterisation of homogeneous fractional {Sobolev} spaces},
  author={Brasco, Lorenzo and G{\'o}mez-Castro, David and V{\'a}zquez, Juan Luis},
  journal={Calculus of Variations and Partial Differential Equations},
  volume={60},
  number={2},
  pages={60},
  year={2021},
  publisher={Springer}
}

@article{ohn2024adaptive,
  title={Adaptive variational {Bayes}: {Optimality}, computation and applications},
  author={Ohn, Ilsang and Lin, Lizhen},
  journal={The Annals of Statistics},
  volume={52},
  number={1},
  pages={335--363},
  year={2024},
  publisher={Institute of Mathematical Statistics}
}

@article{yang2016computational,
  title={ON THE COMPUTATIONAL COMPLEXITY OF HIGH-DIMENSIONAL {Bayesian} VARIABLE SELECTION},
  author={Yang, Yun and Wainwright, Martin J and Jordan, Michael I},
  journal={The Annals of Statistics},
  volume={44},
  number={6},
  pages={2497--2532},
  year={2016}
}

@book{lee1989bayesian,
  title={Bayesian statistics},
  author={Lee, Peter M},
  year={1989},
  publisher={Oxford University Press London}
}

@article{li2024expected,
  title={Expected information gain estimation via density approximations: Sample allocation and dimension reduction},
  author={Li, Fengyi and Baptista, Ricardo and Marzouk, Youssef},
  journal={arXiv preprint arXiv:2411.08390},
  year={2024}
}

@book{raiffa2000applied,
  title={Applied statistical decision theory},
  author={Raiffa, Howard and Schlaifer, Robert},
  year={2000},
  publisher={John Wiley \& Sons}
}

@article{sorscher2022beyond,
  title={Beyond neural scaling laws: {Beating} power law scaling via data pruning},
  author={Sorscher, Ben and Geirhos, Robert and Shekhar, Shashank and Ganguli, Surya and Morcos, Ari},
  journal={Advances in Neural Information Processing Systems},
  volume={35},
  pages={19523--19536},
  year={2022}
}

@article{xie2023data,
  title={Data selection for language models via importance resampling},
  author={Xie, Sang Michael and Santurkar, Shibani and Ma, Tengyu and Liang, Percy S},
  journal={Advances in Neural Information Processing Systems},
  volume={36},
  pages={34201--34227},
  year={2023}
}

@article{rafailov2023direct,
  title={Direct preference optimization: {Your} language model is secretly a reward model},
  author={Rafailov, Rafael and Sharma, Archit and Mitchell, Eric and Manning, Christopher D and Ermon, Stefano and Finn, Chelsea},
  journal={Advances in Neural Information Processing Systems},
  volume={36},
  pages={53728--53741},
  year={2023}
}

@inproceedings{muldrew2024active,
  title={Active Preference Learning for Large Language Models},
  author={Muldrew, William and Hayes, Peter and Zhang, Mingtian and Barber, David},
  booktitle={International Conference on Machine Learning},
  pages={36577--36590},
  year={2024},
  organization={PMLR}
}

@article{wu2023fast,
  title={A fast and scalable computational framework for large-scale high-dimensional {Bayesian} optimal experimental design},
  author={Wu, Keyi and Chen, Peng and Ghattas, Omar},
  journal={SIAM/ASA Journal on Uncertainty Quantification},
  volume={11},
  number={1},
  pages={235--261},
  year={2023},
  publisher={SIAM}
}

@article{chen2021optimal,
  title={Optimal design of acoustic metamaterial cloaks under uncertainty},
  author={Chen, Peng and Haberman, Michael R and Ghattas, Omar},
  journal={Journal of Computational Physics},
  volume={431},
  pages={110114},
  year={2021},
  publisher={Elsevier}
}

@article{birrell2022f,
  title={(f, {$\Gamma$})-Divergences: Interpolating between f-Divergences and Integral Probability Metrics},
  author={Birrell, Jeremiah and Dupuis, Paul and Katsoulakis, Markos A and Pantazis, Yannis and Rey-Bellet, Luc},
  journal={Journal of Machine Learning Research},
  volume={23},
  number={39},
  pages={1--70},
  year={2022}
}

\end{document}